\documentclass{article}

\usepackage[final]{neurips_2025}

\usepackage[utf8]{inputenc} 
\usepackage[T1]{fontenc}    
\usepackage{hyperref}       
\usepackage{url}            
\usepackage{booktabs}       
\usepackage{amsfonts}       
\usepackage{nicefrac}       
\usepackage{microtype}      
\usepackage{xcolor}         
\usepackage{amsmath}
\usepackage{amssymb}
\usepackage{mathtools}
\usepackage{amsthm}
\usepackage{multirow}
\usepackage{color}
\usepackage{pifont}

\usepackage{floatrow}
\newfloatcommand{capbtabbox}{table}[][\FBwidth]
\newcommand{\tablestyle}[2]{\setlength{\tabcolsep}{#1}\renewcommand{\arraystretch}{#2}\centering\small}

\definecolor{brickred}{rgb}{0.8, 0.25, 0.33}
\definecolor{forest}{RGB}{34,139,34}
\newcommand{\methodshorthand}{{TREND}}

\title{TREND: Unsupervised 3D Representation Learning via Temporal Forecasting for LiDAR Perception}

\author{%
Runjian Chen$^{1}$\quad 
Hyoungseob Park$^{2}$\quad 
Bo Zhang$^3$\quad 
Wenqi Shao$^{3}$\quad \\
\textbf{Ping Luo}$^{1,4}$\footnotemark[1]\thanks{Corresponding authors.}\quad
\textbf{Alex Wong}$^{2}$\footnotemark[1]\\[3mm]
$^1$The University of Hong Kong \quad
$^2$Yale University
\\$^3$Shanghai AI Laboratory \quad
$^4$HKU Shanghai Intelligent Computing Research Center
\\
{\tt\small \{rjchen, pluo\}@cs.hku.hk  \quad alex.wong@yale.edu}
}

\begin{document}

\maketitle

\begin{abstract}
Labeling LiDAR point clouds is notoriously time-and-energy-consuming, which spurs recent unsupervised 3D representation learning methods to alleviate the labeling burden in LiDAR perception via pretrained weights. Existing work focus on either masked auto encoding or contrastive learning on LiDAR point clouds, which neglects the temporal LiDAR sequence that naturally accounts for object motion (and their semantics). Instead, we propose \methodshorthand , short for \textbf{T}emporal \textbf{RE}ndering with \textbf{N}eural fiel\textbf{D}, to learn 3D representation via forecasting the future observation in an unsupervised manner. \methodshorthand\ integrates forecasting for 3D pre-training through a Recurrent Embedding scheme to generate 3D embeddings across time and a Temporal LiDAR Neural Field specifically designed for LiDAR modality to represent the 3D scene, with which we compute the loss using differentiable rendering. We evaluate \methodshorthand \ on 3D object detection and LiDAR semantic segmentation tasks on popular datasets, including Once, Waymo, NuScenes, and SemanticKITTI. \methodshorthand\ generally improves from-scratch models across datasets and tasks and brings gains of 1.77\% mAP on Once and 2.11\% mAP on NuScenes, which are up to $400\%$ more improvement compared to previous SOTA unsupervised 3D pre-training methods. Codes and models will be available \href{https://github.com/Runjian-Chen/TREND}{here}.
\end{abstract}

\section{Introduction}
\label{sec:intro}
Light-Detection-And-Ranging (LiDAR) is widely used in autonomous driving. By emitting laser rays into the environment, it provides accurate measurements of the distance along each ray with time-of-flight principle. There has been strong research interest on LiDAR-based perception like 3D object detection \citep{second, centerpoint, pv-rcnn, pv-rcnn++, sst, transfusion, 3d_object_detection_survey} and semantic segmentation \cite{lidar_segmentation_1, lidar_segmentation_2}. However, labeling for LiDAR point clouds is notoriously time-and-energy-consuming. According to \cite{pointcloud_labeling}, it costs an expert labeler at least 10 minutes to label one frame of LiDAR point cloud at a coarse-level and more at finer granularity. Assuming sensor frequency at 20$Hz$, it could cost more than 1000 days of a human expert to annotate a one-hour LiDAR sequence. To alleviate the labeling burden, unsupervised 3D representation learning \cite{pointcontrast, contrastive_scene_context, P4contrast, strl, gcc3d, co3, gdmae, mvjar, ponder, ponderv2, unipad} pre-trains 3D backbone and fine-tune on downstream tasks for performance improvement with the same number of labels.

\begin{figure}
    \centering
    \includegraphics[width=0.98\linewidth]{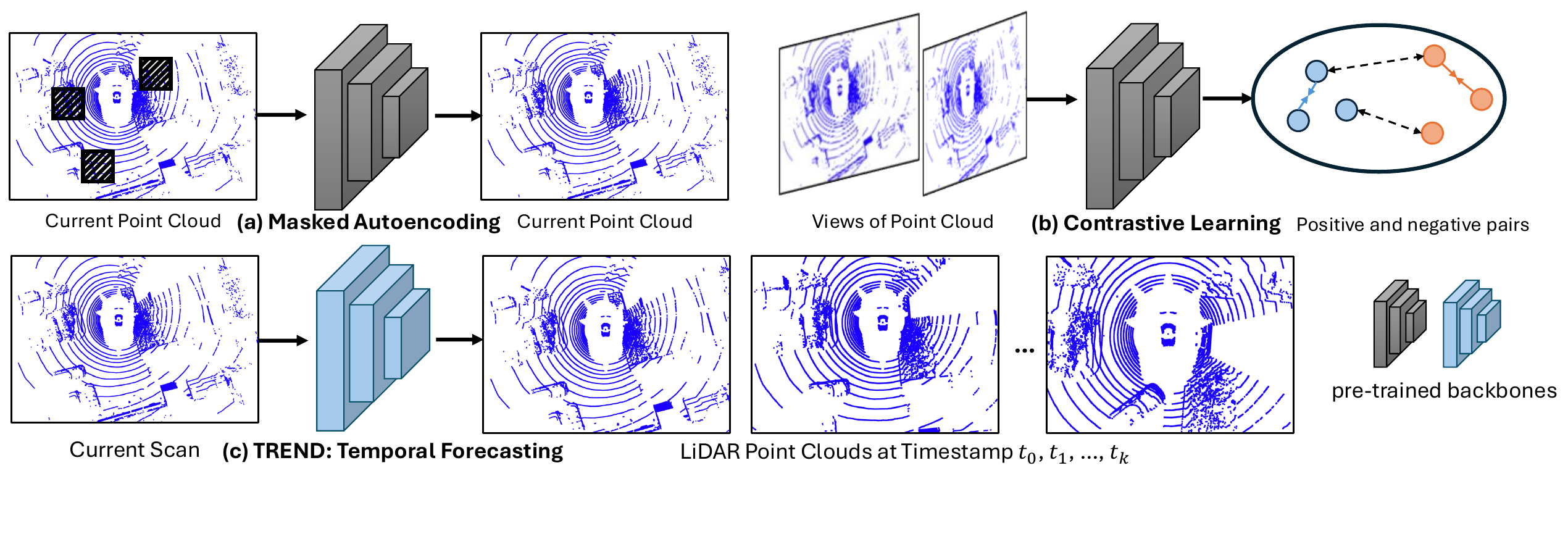}
    \vspace{-3mm}
    \caption{Different schemes for unsupervised 3D representation learning. (a) Masked Autoencoding  applies random masking and pre-train by reconstruction. (b) Contrastive methods build views of point cloud and pre-train by pulling together positive pairs and pushing away negative pairs. (c) \methodshorthand\ explores object motion and semantic information via temporal forecasting in LiDAR sequence.}
    \vspace{-2mm}
\label{fig:teaser}
\end{figure}

Previous literature on unsupervised 3D representation learning for LiDAR perception can be divided into two streams, as shown in Figure \ref{fig:teaser} (a) and (b). (a) Masked-autoencoder-based methods \cite{gdmae, mvjar, ponder, ponderv2, unipad} randomly mask LiDAR point clouds and the pre-training entails reconstructing the masked areas. (b) Contrastive-based methods \cite{gcc3d, co3} construct two views from one frame (or adjacent frames) of LiDAR point cloud and maximize the similarity among positive pairs while minimizing the similarity of negative pairs. Both approaches assume a predefined set of nuisance variability. Nuisance variability refers to variables inherent in the input that should be non-consequential to the outcome, but nonetheless may impact the output. An example of this is orientation: the same object appearing in different orientations can cause the outcome to differ. To obtain the same outcome, one needs to be invariant. In (a), the set of nuisance variability is occlusions, which naturally is induced by motion; in (b) it is the handcrafted set of transformations on LiDAR scans used in contrastive learning. While the procedures are unsupervised, they implicitly select the set of invariants, which benefits the downstream tasks. Unlike them, we subscribe to allowing the data to determine nuisances by simply observing and predicting scene dynamics. This leads to a novel unsupervised 3D representation learning approach based on forecasting LiDAR point clouds (Figure \ref{fig:teaser} (c)). Naturally, points belonging to the same object instance, within a point cloud, tend to move together. By observing current point cloud and predicting future observation, our pre-training scheme implicitly encodes semantics and biases of object interactions over time.

However, leveraging forecasting as unsupervised 3D representation is nontrivial as scene dynamics are often complex and nonlinear. There are two main challenges: 1) How to generate 3D embeddings at different timestamps with current LiDAR scan? 2) How to represent the 3D scene with embeddings and optimize the network via forecasting? 

For 1), there exists tangential work in occupancy prediction field \cite{4docc, uno, copilot4d} that generates 3D features at different timestamps via directly using 3D/2D convolution \cite{4docc, uno} or a diffusion decoder with frozen 3D encoder \cite{copilot4d}. However, actions of the ego-vehicle is not taken into account in \cite{4docc, uno}, which is important for future observation forecasting as the ego actions reflect the interaction between ego-vehicle and other traffic participants. For example, if the ego-vehicle is running at a high speed, pedestrians might stop to avoid accidents. If the ego-vehicle stops at the crossing, pedestrians might start to walk across the road. For 2), applying neural field, as in existing work \cite{ponder, ponderv2, unipad}, to represent the 3D scene at different timestamps yields little to no improvement. The first reason is that the network needs to learn to understand the concept of ``time'' with the 3D convolution, which could be very difficult. The second one is that the neural fields in \cite{ponder, ponderv2, unipad,nerf, neus} are designed for camera modality, which neglects important characteristic in LiDAR modality like intensity. 

We address these challenges by proposing \methodshorthand , short for \textbf{T}emporal \textbf{RE}ndering with \textbf{N}eural fiel\textbf{D}, for unsupervised 3D pre-training. For 1), we propose a Recurrent Embedding scheme, which generates 3D embeddings along time axis with sinusoidal encoding of the ego actions followed by a shallow 3D convolution. This enables us to model ego actions over time, which also assists in forecasting future observations. For 2), we propose a Temporal LiDAR Neural Field that explicit takes timestamps as inputs and integrates LiDAR geometry (surface points) as well as intensity to reconstruct and forecast LiDAR point clouds for optimizing the backbones. While this takes inspiration from existing work in neural field decoders \cite{ponder, ponderv2, unipad,nerf, neus}, it is distinct from them in that our design enables forecasting and also modeling of LiDAR characteristics, such as intensity.

We demonstrate \methodshorthand\ on four benchmark datasets (Once \cite{once}, NuScenes \cite{nuscenes}, Waymo \cite{waymo}, and Semantic Kitti \cite{semantic_kitti}) for the downstream 3D object detection and LiDAR semantic segmentation tasks, where \methodshorthand\ achieves up to $400\%$ more improvement compared to previous SOTA pre-training method for Once (1.77\% mAP) and improves by $90\%$ on NuScenes (2.11\% mAP).
\section{Related Work}
\label{sec:related work}
\noindent\textbf{Pre-training for Point Cloud. }Since annotating 3D point clouds requires significant effort and time, there has been great interest on improving label efficiency for point cloud perception via 3D pre-training. 
For indoor scene point cloud, PointContrast \cite{pointcontrast} first reconstructs the whole scene and uses contrastive learning for pre-training. The research thread was followed by P4Contrast \cite{P4contrast} and Contrastive-Scene-Context \cite{contrastive_scene_context}. For outdoor scene LiDAR point clouds, there exists two primary schools of thought, depending on whether labels are required during the pre-training stage. The first school takes a semi-supervised 3D pre-training approach by utilizing a small sest of labels during pre-training, where the pre-training tasks include object detection, occupancy prediction (e.g., AD-PT \cite{adpt} and SPOT \cite{spot}). The second takes an unsupervised 3D representation learning approach, where no label is required during pre-training: 1) Contrastive-based methods \cite{co3, strl, strl2, gcc3d, triangle_contrast} create alternative (augmented) views of outdoor scene LiDAR point cloud and learns the representation by contrastive learning. 2) Mask-Autoencoder-based methods \cite{gdmae, mvjar, ponder, ponderv2, unipad, tmae}  mask the input LiDAR point clouds and reconstruct the masked elements as the supervision signal. Following these lines of work, \cite{strl, strl2} utilizes adjacent frames of LiDAR point clouds as views for contrastive learning. However, as the scenes are dynamic and there are no labels, the positive and negative pairs selection is very noisy \cite{co3}, resulting in pre-training performance degradation. T-MAE \cite{tmae} proposes to use the adjacent previous frame of LiDAR point clouds for masked autoencoding pre-training, but temporal information is limited to two frames (less than 0.5 second) and only history information is used. Also, ego action is not modeled in \cite{tmae}, which fails to learn the interaction between ego-vehicle and other traffic participants. Furthermore, the decoder in \cite{tmae} is simply Multi-layer Perceptron on occupied 3D space and neglects empty parts of the scenes, which also matters in downstream tasks. For RGB image pretraining, ViDAR \cite{vidar} utilizes future point clouds to pre-train image encoders with occupancy-based decoder, but suffers from the similar problem of \cite{tmae}. Additionally, ViDAR also neglects current LiDAR point clouds, which is important for downstream 3D perception tasks. Besides, some works leverage 2D image prior to pre-train LiDAR encoder \cite{slidr, limoe, sonata}. Unlike them, we propose \methodshorthand\ and use temporal forecasting as the pre-training goal. \methodshorthand\ utilizes a Recurrent Embedding scheme to integrate ego actions for temporal 3D embeddings and a Temporal LiDAR Neural Field as decoder to render both current and future point clouds for pre-training.

\noindent\textbf{Neural Field }plays an important role in 3D scene representation \cite{nerf, neus}. 
IAE \cite{iae} and Ponder \cite{ponder, ponderv2} are pioneering work to use neural field in 3D pre-training and both use reconstruction as pre- training task; UniPAD \cite{unipad} extends this line of work. The neural fields in \cite{nerf, neus, ponder, ponderv2, unipad} are originally designed for camera modality, which neglects characteristic of LiDAR point clouds and temporal information. Unlike them, we explore time-dependent neural field for LiDAR geometry and intensity by proposing a novel pre-training decoder and task to forecast future LiDAR point clouds.

\noindent\textbf{Scene Flow and LiDAR Forecasting. } 3D scene flow \cite{3d_scene_flow_1, 3d_scene_flow_2, 3d_scene_flow_3, 3d_scene_flow_4, 3d_scene_flow_5, 3d_scene_flow_6,3d_scene_flow_7,3d_scene_flow_8,3d_scene_flow_9} has long been explored. Given current and past point clouds, the goal is to estimate per-point translation for the current point clouds. LiDAR forecasting take past and current observations as inputs and predict the future LiDAR point clouds, which necessitates the induction that we hypothesize beneficial for downstream perception tasks. Representative works include 4DOCC \cite{4docc}, Copilot4D \cite{copilot4d} and Uno \cite{uno}. 4DOCC \cite{4docc} uses a U-Net convolutional architecture and conduct differentiable rendering on the BEV feature map to predict the LiDAR observation in the future. Copilot4D \cite{copilot4d} first trains a tokenizer/encoder for LiDAR point cloud with masked-and-reconstruction task and then freeze the encoder to train a diffusion-based decoder for LiDAR forecasting. Uno \cite{uno} proposes to use occupancy field as the scene representation for point cloud forecasting. The forecasting training stage in Copilot4D \cite{copilot4d} does not envolve the 3D encoder for LiDAR point cloud and only focuses on training the diffusion-based decoder, which actually does not introduce temporal information into the 3D encoder. 4DOCC \cite{4docc} and Uno \cite{uno} train the 3D encoder for forecasting but do not take the action of the autonomous vehicle into consideration. However, the interaction between the autonomous vehicle and the traffic participants is important for the prediction. The above methods study treat forecasting as the primary perception task. Unlike them, \methodshorthand\ adopts point cloud forecasting for unsupervised 3D representation learning and aims to improve downstream perception tasks via pre-training. \methodshorthand\ incorporate ego action, which previous works do not account, for pre-training.

\noindent\textbf{LiDAR-based 3D Perception. }LiDAR 3D object detection aims to take the raw LiDAR point clouds as input and predict bounding boxes for different object categories in the scene. Existing literature on LiDAR-based 3D object detection can be divided into three main streams based on the 3D encoder. 1) Point-based methods \cite{pointrcnn, partA2} apply point-level embedding to detect objects in the 3D space. 2) Embraced by \cite{second, centerpoint, transfusion, sst}, voxel-based methods apply voxelization to the raw point clouds and use sparse 3D convolution to encode the 3D voxels. 3) Point-voxel-combination methods \cite{pv-rcnn, pv-rcnn++} combine the point-level and voxel-level features from 1) and 2). LiDAR semantic
segmentation predicts category label for each LiDAR point. Cylinder3D \cite{cylinder3d}, PVKD \cite{pvkd}, Point-Transformer \cite{semantic_segmentation_3,semantic_segmentation_4,semantic_segmentation_5} and SphereFormer \cite{semantic_segmentation_2} achieve excellent performance for LiDAR segmentation task. In this paper, we use both LiDAR 3D object detection and segmentation as downstream tasks and fine-tune on various datasets \cite{once,nuscenes,waymo,semantic_kitti} to evaluate the effectiveness of \methodshorthand.
\section{Method}
\label{sec: method}
\begin{figure}[t]
\centering
\includegraphics[width=0.98\linewidth]{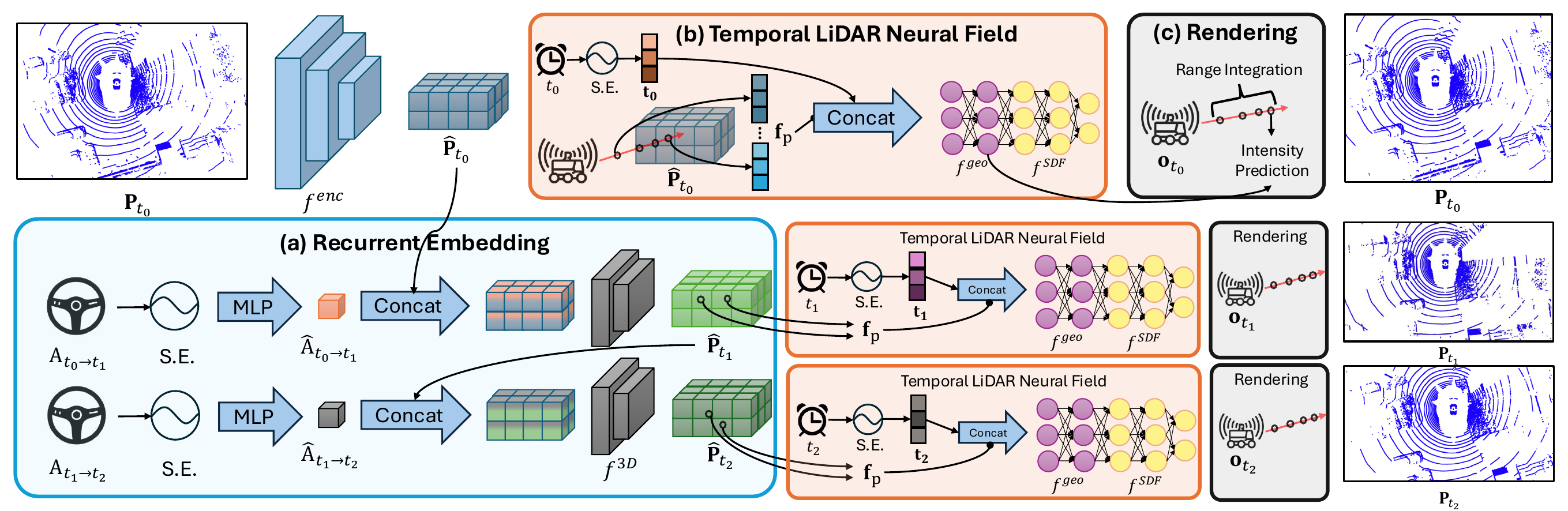} 
\caption{The overview of \methodshorthand. \methodshorthand\ first uses ``S.E.'' (sinusoidal encoding \cite{positional_encoding_1, positional_encoding_2}) and Multi-layer Perceptron to embed ego actions and concatenate them with previous 3D embeddings to generate 3D embeddings at different timestamps (a). Then the Temporal LiDAR Neural Field (b) is used to represent the temporal 3D scene. The queried features, timestamp embeddings and point positions are concatenated and fed into geometry feature function $f^{\text{geo}}$. Next, we separately predict intensity values and signed distance values with geometry features at sampled points and conduct differentiable rendering to reconstruct and forecast LiDAR point clouds for pre-training.}
 \label{figure: pipeline of proposed method}
\end{figure}
In this section, we introduce \methodshorthand\ for unsupervised 3D representation learning on LiDAR perception via temporal forecasting. Fig.~\ref{figure: pipeline of proposed method} is an overview of \methodshorthand. To overcome the two challenges in incorporating temporal forecasting for unsupervised 3D representation learning, we propose (a) the Recurrent Embedding scheme that accounts for the effect of autonomous vehicle's (ego)action to generate 3D embeddings at different timestamps, (b) Temporal LiDAR Neural Field, which represents the 3D scene by the geometry function and signed distance value function. The pre-training goal is to render current and future point clouds to compute loss and optimize the network. We first introduce problem formulation and overall pipeline in Section \ref{subsec: problem formulation and pipeline}. Then we describe the Recurrent Embedding scheme and the Temporal Neural Field in details respectively in Section \ref{subsec: recurrent embedding} and \ref{subsec: lidar neural field}. Finally in Section \ref{subsec: rendering}, we discuss the differentiable rendering process and loss computation.

\subsection{Problem Formulation and Overview}
\label{subsec: problem formulation and pipeline}
\noindent\textbf{Notations.} To start with, LiDAR point clouds are denoted as $\mathbf{P}=[\mathbf{L},\mathbf{F}]\in \mathbb{R}^{N\times (3+d)}$, the concatenation of the $xyz$-location $\mathbf{L}\in \mathbb{R}^{N\times 3}$ and point features $\mathbf{F}\in \mathbb{R}^{N\times d}$. Here $N$ denotes the number of points in the point clouds and $d$ the number of feature channels. For instance, $d=1$ in Once \cite{once} representing intensity and for Waymo \cite{waymo}, $d=2$ are intensity and elongation. To indicate point clouds at different timestamps, we use subscripts and $\mathbf{P}_{t}=[\mathbf{L}_{t},\mathbf{F}_{t}]\in\mathbb{R}^{N_{t}\times (3+d)}$ is point cloud at time $t\in\{t_0,t_1,t_2,...,t_k\}$, where $t_0$ indicates current timestamp and $t_1,t_2,...t_k$ are future timestamps. At each timestamp $t_n$, we also have the action $\mathbf{A}_{t_n\rightarrow t_{n+1}}=[\Delta_x,\Delta_y,\Delta_{\theta}]\in\mathbb{R}^3$ of the autonomous vehicle and it is described with the relative translation on x-y plane ($\Delta_x,\Delta_y$) and orientation with respect to z-axis ($\Delta_{\theta}$) between timestamp $t_n$ and $t_{n+1}$.

\noindent\textbf{Overview.} Our goal is to pre-train the 3D encoder $f^{\text{enc}}$ in an unsupervised manner via forecasting. We begin by embedding $\mathbf{P}_{t_0}$ with the 3D encoder $f^{\text{enc}}$ to obtain the 3D representations
\begin{equation}
\hat{\mathbf{P}}_{t_0}=f^{\text{enc}}(\mathbf{P}_{t_0}),
\end{equation}
where $\hat{\mathbf{P}}_{t_0}\in\mathbb{R}^{D\times H\times W\times \hat{d}}$ denotes the embedded 3D features with spatial resolution of $D\times H\times W$ and $\hat{d}$ feature channels. Then with $\hat{\mathbf{P}}_{t_0}$ and action at different timestamps $\mathbf{A}_{t_n\rightarrow t_{n+1}}$ as inputs, we apply the recurrent embedding scheme $f^{\text{rec}}$ and get the 3D embedding at different timestamps
\begin{equation}
\hat{\mathbf{P}}_{t_{n+1}}=f^{\text{rec}}(\mathbf{A}_{t_n\rightarrow t_{n+1}},\hat{\mathbf{P}}_{t_n}),
\end{equation}
where $n=0,1,...$. Finally, to guide the training of 3D encoder in an unsupervised manner, we use a Temporal Neural Field to reconstruct and forecast LiDAR point clouds $\Tilde{\mathbf{P}}_{t_n}$
\begin{equation}
\Tilde{\mathbf{P}}_{t_n}=f^{\text{render}}(\hat{\mathbf{P}}_{t_n}),
\end{equation}
and compute the loss against the raw observation $\mathbf{P}_{t_n}$ for optimization. Note that all the LiDAR point clouds are transformed into the coordinate frame of $t_0$ for consistency.

\subsection{Recurrent Embedding Scheme}
\label{subsec: recurrent embedding}
In order to introduce temporal information into 3D pre-training for $f^{\text{enc}}$, we embed the 3D representation $\mathbf{P}_{t_0}$ of the current frame at $t_0$ into future 3D representation ($\mathbf{P}_{t_1}$, $\mathbf{P}_{t_2}$ ...). To achieve this, previous literature \cite{4docc, uno} directly apply learnable 3D/2D decoders but neglect the effect of autonomous vehicle's action $\mathbf{A}_{t_n\rightarrow t_{n+1}}$. However, the action of the autonomous vehicle is a part of the interaction between the ego-vehicle and traffic participants, and may influence the motion of the traffic participants on the road; hence, serving as a predictor. For example, if the autonomous vehicle does not move for some time, other traffic participants might move faster and vice versa. Thus, we propose to take $\mathbf{A}_{t_n\rightarrow t_{n+1}}$ into account and use a recurrent embedding scheme.

To begin, sinusoidal encoding \cite{positional_encoding_1, positional_encoding_2} are used to encode the relative translational components $[\Delta_x, \Delta_y]$ in raw action $\mathbf{A}_{t_n\rightarrow t_{n+1}}$ with sinusoidal functions of different frequencies. The resulting translation feature $\mathbf{f}_{\text{tl}}\in\mathbb{R}^{d_{\text{sin}}}$ contains $d_{\text{sin}}$ bounded scalars. Then we use $\mathbf{f}_{\text{rot}}=[\sin{\Delta_\theta},\cos{\Delta_\theta}]\in\mathbb{R}^2$ to represent the rotational component in $\mathbf{A}_{t_n\rightarrow t_{n+1}}$ and concatenate both features to generate an initial action embedding $\Tilde{\mathbf{A}}_{t_n\rightarrow t_{n+1}}=[\mathbf{f}_{\text{tl}}, \mathbf{f}_{\text{rot}}]\in\mathbb{R}^{d_{\text{sin}}+2}$ for $\mathbf{A}_{t_n\rightarrow t_{n+1}}$. Note, this initial embedding process does not require any learnable parameter. To then further learn to embed $\Tilde{\mathbf{A}}_{t_n\rightarrow t_{n+1}}$, we apply a shared shallow multi-layer perceptron (MLP) $f^{\text{act}}$ and project it to $\hat{\mathbf{A}}_{t_n\rightarrow t_{n+1}}\in\mathbb{R}^{d_{\text{act}}}$
\begin{equation}
\hat{\mathbf{A}}_{t_n\rightarrow t_{n+1}} = f^{\text{act}}(\Tilde{\mathbf{A}}_{t_n\rightarrow t_{n+1}}).
\end{equation}
With 3D embeddings at current timestamp $\hat{\mathbf{P}}_{t_0}$ and action embeddings at future timestamps $\hat{\mathbf{A}}_{t_n\rightarrow t_{n+1}}$, we broadcast $\hat{\mathbf{A}}_{t_n\rightarrow t_{n+1}}$ to the shape of $\hat{\mathbf{P}}_{t_n}$ and concatenate it with $\hat{\mathbf{P}}_{t_n}$ along the  feature dimension, followed by a shared shallow 3D dense convolution $f^{\text{3D}}$ to get the embedding at different timestamps $\hat{\mathbf{P}}_{t_{n+1}}\in\mathbb{R}^{D\times H\times W\times \hat{d}}$.
\begin{equation}
\hat{\mathbf{P}}_{t_{n+1}}=f^{\text{3D}}([\mathbf{A}_{t_n\rightarrow t_{n+1}},\hat{\mathbf{P}}_{t_n}]),\ n=0,1,...
\end{equation}
While local features reflect the understanding of other traffic participants and the environment, the concatenation provides local features with understanding of ego-vehicle motion. Despite the feature vector containing the vehicle ego-motion, the remainder of the feature vector allows us to predict the feature evolution. This recurrent embedding scheme allows us to model the evolution of the latent scene features based on vehicle ego-motion.

\subsection{Temporal LiDAR Neural Field}
\label{subsec: lidar neural field}
We propose to use Neural field to represent the 3D scene around the autonomous vehicle at different timestamp $t$, which is the basis for LiDAR point clouds rendering. Previous work \cite{nerf, neus, lidar_nerf_1, lidar_nerf_2, lidar_nerf_3} design neural field for image modality and neglect both LiDAR characteristic and temporal information. On the contrary, we propose Temporal LiDAR Neural Field. As shown in Fig.~\ref{figure: pipeline of proposed method}, the goal of Temporal LiDAR Neural Field is to infer the geometry features and the signed distance value \cite{sdf_1, sdf_2} for a point $\mathbf{p}$ in 3D space at timestamp $t$. Given the location of a specific point $\mathbf{p}=[x,y,z]\in\mathbb{R}^3$ at timestamp $t$, we first query the feature $\mathbf{f}_{\text{p}}\in\mathbb{R}^{\hat{d}}$ at $\mathbf{p}$ with $\hat{\mathbf{P}}_t$ by trilinear interpolation $f^{\text{tri}}$ implemented in Pytorch \cite{pytorch}:
\begin{equation}
    \mathbf{f}_{\text{p}} = f^{\text{tri}}(\mathbf{p}, \hat{\mathbf{P}}_t).
\end{equation}
Similar to initial action embedding in Section \ref{subsec: recurrent embedding}, we apply sinusoidal encoding \cite{positional_encoding_1, positional_encoding_2} to encode timestamp $t$ to $\mathbf{f}_{\text{t}}\in\mathbb{R}^{d_{\text{sin}}}$. Taking the concatenation of location $\mathbf{p}$, $\mathbf{f}_{\text{t}}$ and the queried feature $\mathbf{f}_{\text{p}}$ as inputs, we first predict the geometry features $\mathbf{f}_{\text{geo}}\in\mathbb{R}^{d_{\text{geo}}}$ with $f^{\text{geo}}$ and then the signed distance value $s\in\mathbb{R}$ \cite{sdf_1, sdf_2} with $f^{\text{SDF}}$, which are parameterized by Multi-layer Perceptron:
\begin{equation}
\mathbf{f}_{\text{geo}} = f^{\text{geo}}([\mathbf{p}, \mathbf{t}, \mathbf{f}])\quad;\quad
s = f^{\text{SDF}}(\mathbf{f}_{\text{geo}}).
\end{equation} 

\subsection{Point Cloud Rendering}
\label{subsec: rendering}
Each LiDAR point $\mathbf{p}$ can described by the sensor origin $\mathbf{o}\in\mathbb{R}^3$, normalized direction $\mathbf{d}\in\mathbb{R}^3$, and the range $r\in\mathbb{R}$, i.e., $\mathbf{p}=o+r\mathbf{d}$. Similar to \cite{nerf, neus, lidar_nerf_1, lidar_nerf_2, lidar_nerf_3}, we first sample $N_{\text{render}}$ rays at the sensor position $\mathbf{o}$ that trave along the normalized direction $\mathbf{d}$, and apply differentiable rendering to predict the range of LiDAR beam rays at different timestamp $t\in\{t_0,t_1,t_2,...\}$ with our Temporal Neural Field.

\noindent\textbf{Sampling of $N_{\text{render}}$. }LiDAR points on the ground are less informative and we filter out ground points by setting a threshold $z_{\text{thd}}$ for $z$ values of the point position in vehicle coordinate frame. $z_{\text{thd}}$ is determined by sensor height provided in the datasets. After that, we uniformly sample $N_{\text{render}}$ at timestamp $t_n$ to conduct range rendering and loss computation.


\noindent\textbf{Range Rendering. }For a specific timestamp $t$, we sample $N_{\text{ray}}$ points following \cite{neus} along each ray and construct the point set $\{\mathbf{p}_n=\mathbf{o}+r_n\mathbf{d}\}_{n=1}^{N_{\text{ray}}}$. For each point in the point set, we estimate the signed distance value $s_n$ as described in Section \ref{subsec: lidar neural field}. Then we predict the occupancy value $\alpha_n$
\begin{equation}
    \alpha_n = \max{(\frac{\Phi_z(s_n)-\Phi_z(s_{n+1})}{\Phi_z(s_n)},0)},
\end{equation}
where $\Phi_z(x)=(1+e^{-zx})^{-1}$ is the sigmoid function with a learnable scalar $z$. With $\alpha_n$, we estimate the accumulated transmittance $\mathcal{T}_n$ \cite{neus} by $\mathcal{T}_n = \prod^{n-1}_{i=1}(1-\alpha_i)$. 
We follow conventional rendering methods \cite{neus} to compute an occlusion-aware and unbiased weight $w_n= \mathcal{T}_n \alpha_n$. 
Differentiable rendering is conducted by integrating sampled points along the ray, leading to the predicted range $\Tilde{r}$, 
\begin{equation}
\Tilde{r}=\sum_{n=1}^{N_{\text{ray}}}w_n*r_n.
\end{equation}

\noindent\textbf{Intensity Prediction. }According to \cite{intensity_prediction}, the intensity of LiDAR point clouds is decided by three factors: sensor system, surface material, and injection angle. Moreover, injection angle can be inferred by the ray direction $\mathbf{d}$ and surface normal. The geometry feature $\mathbf{f}_{\text{geo}}$ and queried feature $\mathbf{f}_{p}$ at the scanned point includes information about the surface normal and material respectively. Thus we first embed the ray direction $\mathbf{d}$ by a Multi-layer Perceptron $f^{\text{dir}}$. Then we concatenate the direction embedding $\mathbf{f}_{\text{dir}}\in\mathbb{R}^{d_{\text{dir}}}$, geometry feature $\mathbf{f}_{\text{geo}}$ and queried feature $\mathbf{f}_{p}$ at the scanned point and apply an intensity network $f^{\text{int}}$ to predict the intensity $\Tilde{\mathcal{I}}$
\begin{equation}
   \Tilde{\mathcal{I}} = f^{\text{int}}([\mathbf{f}_{\text{dir}}, \mathbf{f}_{\text{geo}}, \mathbf{f}_{p}]).
\end{equation}

\noindent\textbf{Loss Function. }For each sampled ray, we have the observed range $r^i$ and intensity $\mathcal{I}^i$ and the predicted ones $\Tilde{r}^i$ and $\Tilde{\mathcal{I}}^i$, with which we compute an L1 loss; meanwhile, the expected signed distance value of the observed points $s_i$ is zero. We integrate this constraint into the loss function. 
\begin{equation}
    \mathcal{L}_{t_n} = \frac{1}{N_{\text{render}}}\sum_{i=1}^{N_{\text{render}}}(|r^i-\Tilde{r}^i| + |\mathcal{I}^i-\Tilde{\mathcal{I}}^i|+|s_i|).
\end{equation}

\subsection{Curriculum Learning for Forecasting Length}
It is difficult for a randomly initialized network to directly learn to forecast several frames of LiDAR point clouds. Thus we propose to borrow the idea of curriculum learning \cite{curriculum_learning_1, curriculum_learning_2} and gradually increase the forecasting length. Specifically, we optimize the network with $N_{\text{curri}}^{l}$ curriculum learning epochs for $\{\mathbf{P}_{t_n}\}_{n=0}^{l}$, where $l=1,2,...$. Because the observation nearer to current timestamp introduce more information about the current stage, we always reconstruct the current LiDAR point clouds and apply a decay weights $p(m)$ ($m=1,2,...,l$) to sample a future timestamp, where $p(m)>p(m+1)$ always holds. The final loss is computed as,
\begin{equation}
    \mathcal{L} = \mathcal{L}_{t_0} + \mathcal{L}_{t_m},\ \ m\sim p(m).
\end{equation}

\subsection{Discussions}
\label{discussions}

\noindent\textbf{Theoretical Insight of \methodshorthand. }We provide an analysis in the aspect of information theory \cite{information_bottleneck_1} and minimal sufficient representation \cite{information_bottleneck_2, information_bottleneck_3, information_bottleneck_4}. Let data be $\mathbf{X}$, its representation be  $\mathbf{Z}$ and a downstream task be  $\mathbf{Y}$.  $\mathbf{Z}$ is sufficient for  $\mathbf{Y}$ if it is faithful to the task, e.g., fidelity of predictions. However, one may choose a  $\mathbf{Z}$, including $\mathbf{X}$ -- by the definition of data processing inequality, if  $\mathbf{X}$ is sufficient, then  $\mathbf{Z}$ is also sufficient. According to our discussion in Section \ref{sec:intro}, there are factors (nuisances) in the data that (negatively) impact predictions, and has implications towards generalization. Hence, it is desirable for a representation to be minimal, that is, containing the smallest amount of information, but sufficient for $\mathbf{Y}$. The instantiation of this is the Information Bottleneck (IB) Lagrangian:
\begin{equation}
\max\ I(\mathbf{Z}; \mathbf{Y}) - \beta I(\mathbf{X}; \mathbf{Z}),
\end{equation}
where $I(;)$ denotes the mutual information between two random variables. Maximizing IB Lagrangian leads to fidelity for the task through the first data term and minimality or compression through the second bottleneck term. Naturally, $\beta$ controls the compression, where larger compression discards nuisance variability. The nuisance $\mathbf{N}$ influence $\mathbf{Z}$ only through $\mathbf{X}$, which follows the casual chain $\mathbf{N}$ --> $\mathbf{X}$ --> $\mathbf{Z}$. 
Thus we have $I(\mathbf{Z}; \mathbf{N}) \leq I(\mathbf{X}; \mathbf{Z}) - I(\mathbf{Z}; \mathbf{Y})$. Hence, the relationship between IB Langrangian and our proposal of temporal forecasting as a mechanism for unsupervised representation learning lies in the choice of modeling nuisance variables $\mathbf{N}$. What we want to accomplish is to minimize $I(\mathbf{Z}; \mathbf{N})$. We posit that the temporal dynamics within a dataset better exhibit the set of nuisance variables than does a handcrafted set through data augmentation. While it is intractable to quantify $I(\mathbf{Z}; \mathbf{N})$ directly, our empirical findings suggest that representations learned through temporal forecasting better suppress nuisances and improve downstream performance.

\noindent\textbf{Memory and Computational Overhead of \methodshorthand. }While TREND introduces temporal forecasting and neural field rendering, the actual memory costs are comparable to baseline methods. We utilize two design choices when sampling the rendering rays: ground point filtering and uniform ray sampling to make the GPU memory consumption feasible for \methodshorthand. In our experiments, all pre-training methods utilize the same GPU memory. As for computational cost during pre-training, since TREND employs Recurrent Embedding scheme, TREND requires approximately 8\% more time than previous methods per epoch (65 mins v.s. 60 mins on 8-A100) for pre-training. Besides, recurrent embedding and temporal neural fields are not used during both fine-tuning and inference. The downstream model architecture, computational cost, and memory usage are identical across all methods.
\section{Experiments}
\label{sec: exps}
Unsupervised 3D representation learning aims to pre-train 3D backbones and use the pre-trained weights to initialize downstream models for performance improvement. In this section, we design experiments to demonstrate the effectiveness of \methodshorthand\ as compared to previous methods. We start with introducing experiment settings in Section \ref{subsec: exp settings}. Then main results are provided in Section \ref{subsec: main results}. Finally, additional experiment results and ablation study are discussed in Section \ref{subsec: other results}.
\vspace{-1mm}

\subsection{Experiment Settings}
\label{subsec: exp settings}
\noindent\textbf{Datasets. }We conduct experiments on four popular autonomous driving datasets including Once \cite{once} NuScenes \cite{nuscenes}, Waymo \cite{waymo} and SemanticKITTI \cite{semantic_kitti}. Once utilizes a 40-beam LiDAR to collect 1 million LiDAR frames and labels 15k of them. Due to the computation resource limitation, we conduct pre-training with \methodshorthand\ on the small split of the unlabeled data (100k frames) and fine-tune the pre-trained backbone with the labeled training set. NuScenes uses a 32-beam LiDAR to collect 1000 scenes in Boston and Singapore, where 850 of them are used for training and the other 150 ones for validation. We use the whole training set without label for all the pre-training methods. Waymo equips the autonomous vehicle with one top 64-beam LiDAR and 4 corner LiDARs to collect point clouds. We use Waymo for evaluating the transferring ability of \methodshorthand. SemanticKITTI uses a 64-beam LiDAR for data collection and provides semantic labels for each point.

\begin{table}[t]
\setlength{\tabcolsep}{2.6pt}
\renewcommand\arraystretch{1.1}
\centering
\begin{tabular}{c|c|c|ccc|ccc|ccc}
\hline
\multirow{2}{*}{Init.} & \multirow{2}{*}{F.T.}  & \multirow{2}{*}{mAP} & \multicolumn{3}{c|}{Vehicle} & \multicolumn{3}{c|}{Pedestrian} & \multicolumn{3}{c}{Cyclist} \\
                       &                        &                      & 0-30m   & 30-50m   & 50m-    & 0-30m    & 30-50m    & 50m-     & 0-30m   & 30-50m   & 50m-   \\ \hline
Ran.                   &          \multirow{7}{*}{5\%}              & 46.07                & 76.71   & 51.15    & 31.84   & 37.53    & 20.12     & 9.84     & 62.00   & 42.61    & 24.18  \\
\cite{ALSO}                   &                        &  44.69 $^{\bf\color{brickred}{-1.38}}$           &  74.04  &  49.66   &  29.63  & 33.98  &  20.94    &   12.42   & 60.63   &  43.14   & 23.63  \\
\cite{occ_mae}                  &                        &  44.43     $^{\bf\color{brickred}{-1.64}}$         &  76.52  &  49.48   &  30.18  &  35.32 &   18.96   &  9.36    &  60.47  &   40.94  &  22.99 \\
 \cite{4docc}                  &                        &    40.84   $^{\bf\color{brickred}{-5.23}}$         &  74.23  &  46.64   &  29.45  & 29.85  &  17.31    &   9.56   &  57.47  &  33.59   &   18.34 \\
\cite{tmae}                   &                        &   45.12 $^{\bf\color{brickred}{-0.95}}$             &  74.20  &  49.52   & 30.25   & 37.51  &   20.46   &  9.97    &  60.93  &   41.82  &  25.75 \\
 \cite{unipad}                  &                        & 46.23   $^{\bf\color{forest}{+0.16}}$              &  78.76  & 55.77    &  37.81  & 31.65  & 16.09     & 8.78     &  64.90  &  44.18   & 24.73  \\
Ours                          &                        &          \bf47.84 $^{\bf\color{forest}{+1.77}}$                   & 79.14   & 55.68    & 36.34   & 35.23    & 18.00     & 11.18    & 64.99   & 45.80    & 28.15  \\ \hline
Ran.                   &         \multirow{7}{*}{20\%}                & 57.68                & 82.70   & 63.37    & 46.34   & 52.61    & 36.48     & 19.03    & 71.03   & 55.34    & 36.34  \\
\cite{ALSO}                   &                        &    56.27    $^{\bf\color{brickred}{-1.41}}$         &  81.01   &  61.13   & 43.63   & 49.78  & 35.51     &  20.02    &   69.55 & 52.58    & 34.94  \\ \cite{occ_mae}                  &                        &  57.09       $^{\bf\color{brickred}{-0.59}}$       &  83.51  & 62.57    &  46.28  & 50.96  &   34.55   &  17.90    &  70.37  & 54.50    & 36.79  \\
\cite{4docc}                   &                        &   54.30      $^{\bf\color{brickred}{-3.38}}$       & 80.69   & 58.95    & 42.13   & 45.09  & 33.14     &  18.04    & 68.90   &  52.20   &  35.09 \\
\cite{tmae}                  &                        &   57.23     $^{\bf\color{brickred}{-0.45}}$        & 81.66   & 62.64    & 45.14   & 51.32  &  34.80    &  17.26    & 70.87   &  54.08   & 33.25  \\
\cite{unipad}                 &                        &    58.08$^{\bf\color{forest}{+0.40}}$           & 84.23   & 65.44    & 48.65   & 49.48  &  34.84    & 19.38     & 70.76   &  55.75   & 38.89  \\
Ours                          &                        &        \bf58.93 $^{\bf\color{forest}{+1.25}}$                   & 84.08   & 65.80    & 50.51   & 50.31    & 33.37     & 19.42    & 72.54   & 56.31    & 39.26  \\ \hline
Ran.                   &         \multirow{7}{*}{100\%}                & 65.03                & 88.18   & 74.23    & 61.75   & 57.32    & 38.90     & 21.96    & 78.07   & 64.32    & 48.16  \\
\cite{ALSO}         &           &      64.19    $^{\bf\color{brickred}{-0.84}}$             &   86.07             &  72.44  & 59.28    & 57.25   & 37.14  &  22.25   & 77.62    &  61.94  &    45.91  \\
\cite{occ_mae}                  &                        &  65.10   $^{\bf\color{forest}{+0.07}}$           & 88.02   &  74.01   &  61.95  & 57.56  &   38.43   &  22.45    &  79.95  &   63.64  &  47.89 \\
 \cite{4docc}                  &                        &   64.48     $^{\bf\color{brickred}{-0.55}}$         &  88.34  &  74.20   &  61.32  & 55.78  &  37.14    &   22.32   &  77.95  &  62.42   & 46.40  \\
\cite{tmae}                &                        &     65.25   $^{\bf\color{forest}{+0.22}}$        &  88.31  &   72.67  &  62.87  & 57.48  &   39.55   &  24.30    &  77.92  &  63.07   & 48.34  \\
\cite{unipad}                 &                        &    65.19    $^{\bf\color{forest}{+0.16}}$        &  88.11  & 74.00    & 62.28   & 57.67  &  38.49    &  21.99    &  79.51  & 64.40    & 47.65  \\
Ours                          &                        &       \bf66.09 $^{\bf\color{forest}{+1.06}}$                & 88.56   & 75.02    & 63.10   & 57.83    & 39.29     & 20.63    & 79.48   & 65.08    & 49.02  \\ \hline
\end{tabular}
\caption{Results on Once dataset \cite{once}. ``Init.'' indicates the initialization methods and ``Ran.'' is random initialization. ``F.T.'' is the ratio of sampled training data for fine-tuning stage. We show mAP for overall performance and APs for different categories within different ranges. Green color is used to highlight the performance improvement and red one for degradation. We also use bold font to highlight the best mAP at different fine-tuning ratio. All the results are in \%.}
\label{table: once fine-tuning}
\vspace{-1mm}
\end{table}

\noindent\textbf{Downstream Models and Evaluation Metrics. }We perform downstream 3D object detection task on Once \cite{once}, NuScenes \cite{nuscenes} and Waymo \cite{waymo} and LiDAR semantic segmentation task on SemanticKITTI \cite{semantic_kitti}. We follow the implementations in the popular code repository called OpenPCDet \cite{pcdet} and select the SOTA models on different datasets. For Once and Waymo, we use CenterPoint \cite{centerpoint} as the downstream model. For Once, Average precisions for different categories within different ranges (APs) and mean average precision (mAP) are used for evaluation. For Waymo, APs and APs with heading (APHs) computed at two difficulty levels (Level-1 and Level-2)  are utilized. For NuScenes, we use Transfusion-LiDAR \cite{transfusion} as the downstream model. APs for different categories, mAP and NuScenes Detection Score (NDS) are used for evaluation. For SemanticKITTI, We use Cylinder3D \cite{cylinder3d} and Mean Intersection over Union (mIoU) and accuracy are computed.

\noindent\textbf{Downstream Training Setting. }The main goal of unsupervised 3D pre-training is to improve \textit{\textbf{sample efficiency instead of accelerating convergence}}, which has been discussed in previous literature \cite{kaiming_rethinking, mvjar}. Sample efficiency means the best performance we can achieve with the same model and the same number of labeled data. Thus, we first gradually increase the training iterations for randomly initialized models until convergence is observed, which means increasing number of training iterations does not further improve the performance. Then we fix the training iterations and use the same schedule for downstream fine-tuning with different pre-training methods. 


\noindent\textbf{Baseline 3D Pre-training Methods. }
We select five baseline methods. (1) ALSO \cite{ALSO}, an occupancy-based method. (2) Occupancy-MAE \cite{occ_mae}, an masked-autoencoder method. (3) 4DOCC \cite{4docc}, a LiDAR point cloud forecasting method. (4) T-MAE \cite{tmae}, a concurrent work that utilizes previous adjacent frame of LiDAR point clouds for mased-and-reconstruction without considering action of the autonomous vehicle. (5) UniPAD \cite{unipad}, the masked-and-reconstruction-based method with rendering decoder. All the pre-trainings for baseline methods are conducted with their official code.

\noindent\textbf{Implementation.} For $f^{\text{enc}}$, we select the popular sparse convolution backbone \cite{submanifold_spconv}. The feature channels for embedded 3D feaures $\hat{\mathbf{P}_{t_n}}$, sinusoidal encoding and action embeddings are respectively set to $\hat{d}=128$, $d_{\text{sin}}=32$ and $d_{\text{act}}=16$. The sampled ray number for rendering is $N_{\text{render}}=12288$ and number of points along each ray is $N_{\text{ray}}=48$. We set the pre-training learning rate as 0.0002 with a cosine learning schedule. \textbf{Random seed is fixed} for all pre-training and fine-tuning to guarantee reproducibility. More details can be found in Appendix \ref{appendix: implementation details}.

\vspace{-2mm}
\subsection{Main Results}
\label{subsec: main results}
\vspace{-1mm}

\noindent\textbf{Results on Once Dataset. }We pre-train \methodshorthand\ and baseline methods on the small split of unlabeled data in Once and uniformly sample $5\%$, $20\%$ and $100\%$ of the labeled training set for downstream fine-tuning. The results are shown in Table \ref{table: once fine-tuning}. For overall performance, it can be found that \methodshorthand\ achieves the best performance across different ratio of fine-tuning data. The performance improvement compared to train-from-scratch model is 1.77, 1.25 and 1.06 respectively for $5\%$, $20\%$ and $100\%$ fine-tuning data, which is up to 4 times more than previous 3D unsupervised pre-training methods and demonstrates the effectiveness of \methodshorthand. As for different categories, \methodshorthand\ achieves up to 4\% mAP improvement on Vehicle and Cyclist for $5\%$ fine-tuning data and generally improve these two categories within different ranges. It can also be found that for Pedestrian class, \methodshorthand\ improves with $100\%$ downstream data but degrades the performance a little bit under $5\%$ and $20\%$ fine-tuning data settings. We think this is because LiDAR point clouds stand for geometry and pedestrians are always captured in LiDAR point clouds with a cylinder-like shape, which is less-distinguishable as compared to cyclists and vehicle. For example, trash bins or poles also appear to be cylinder-like in LiDAR. Thus learning to reconstruct and forecast such less-distinguishable geometry harms the ability of the pre-trained backbone to identify pedestrians among similar cylinder-like shapes especially when there are less labeled downstream data, leading to a little degradation for $5\%$ and $20\%$ settings. Similar phenomenon is also observed for differentiable reconstruction method UniPAD.


\begin{table}[t]
\centering
\footnotesize{
\renewcommand\arraystretch{1.1}
\begin{tabular}{ccccccccccc}
\hline
Init. & mAP   & NDS   & Car   & Truck  & Bus  & Bar. & Mot. & Bic. & Ped. & T.C. \\ \hline
Ran. & 31.06 & 44.75 & 69.18      & 28.73                          &    34.57        &    42.31     &   13.72         &    8.72     &     69.18       &      41.14        \\
\cite{ALSO}    & 30.14 $^{\bf\color{brickred}{-0.92}}$  &   43.73 $^{\bf\color{brickred}{-1.02}}$    &       66.89                 &    25.67   &   34.36   &43.06   &   12.98    &    7.1      &          66.28     &     41.63        \\
\cite{occ_mae}   & 29.94 $^{\bf\color{brickred}{-1.12}}$   &   43.93 $^{\bf\color{brickred}{-0.82}}$   &     68.51                   &    26.32   & 30.90        &   41.74    &     12.36     &       7.0        &     67.84    &  41.27  \\
\cite{4docc}   & 26.99 $^{\bf\color{brickred}{-4.07}}$  & 40.97 $^{\bf\color{brickred}{-3.78}}$  &  67.44     &  25.40                       &  29.37     &    35.58     &    9.53     &  5.16          &      65.26           &       29.47       \\
\cite{tmae}   &        30.53  $^{\bf\color{brickred}{-0.53}}$        &  44.55 $^{\bf\color{brickred}{-0.20}}$     &   68.63                   &     26.02  &     34.66    &   43.98      &  13.21          &    7.26     &     68.78       &      39.82        \\
\cite{unipad}  & 32.16 $^{\bf\color{forest}{+1.10}}$ & 45.50 $^{\bf\color{forest}{+0.75}}$ &    69.82   &    29.54                    &  35.73          &    46.79     &  13.65          &     7.98    &      70.45      &    42.73         \\

Ours  & \bf33.17 $^{\bf\color{forest}{+2.11}}$ & \bf46.21 $^{\bf\color{forest}{+1.46}}$ & 71.24 & 30.08                & 39.57     & 45.42   & 16.65      & 9.33    & 71.84      & 43.70        \\ \hline
\end{tabular}
}
\vspace{-1mm}
\caption{Results on NuScenes \cite{nuscenes} dataset. ``Init.'' indicates the initialization methods and ``Ran.'' is random initialization. We use green color to highlight performance improvement and red for degradation and bold fonts for best performance in mAP and NDS. All the results are in \%.}
\label{table: nuscenes few-shot results}
\end{table}

\begin{table}
\vspace{-1mm}
\tablestyle{5pt}{1.1}
\begin{floatrow}
\capbtabbox{
\begin{tabular}{c|ccccc}
\hline
\multirow{2}{*}{Init.} & \multicolumn{2}{c}{Level-1} & \multicolumn{2}{c}{Level-2} & \multirow{2}{*}{$\bar\Delta$} \\
                       & mAP          & mAPH         & mAP          & mAPH         &                \\ \hline
Ran.                   &      61.60        &   58.58           &      55.62        &     52.87         &                   \\
\cite{unipad}                       &      61.57        &       58.57       &       55.60       &   52.83           &       \bf\color{brickred}{-0.03}            \\
Ours                   &      62.32        &  59.22            &    56.37          &     53.84         &           \bf\color{forest}{+0.77}        \\ \hline
\end{tabular}
}{
 \caption{Results for transferring experiments.}
 \label{table: transferring}
}
\capbtabbox{
\begin{tabular}{ccccc}
\hline
Init. & mAP & Veh. & Ped. & Cyc. \\ \hline
 Ran.*     &  20.48   & 37.88     & 10.62     & 12.96     \\
\cite{tmae}*      &  21.58 $^{\bf\color{forest}{+1.10}}$   &  37.83    &  11.72    & 15.19     \\
\cite{unipad}*     &   24.41 $^{\bf\color{forest}{+3.93}}$  &  40.66    & 12.01     &  20.55    \\
 Ours*     &   29.95 $^{\bf\color{forest}{+9.47}}$  &  44.99    & 16.28     & 28.59     \\ \hline
\end{tabular}
}{
 \caption{Results for accelerating convergence.}
 \label{table: once acceleration}
 \small
}
\end{floatrow}
\vspace{-4mm}
\end{table}

\noindent\textbf{Results on NuScenes Dataset. }We pre-train \methodshorthand\ and baseline methods on the whole training set of NuScenes dataset. We then uniformly sample 175 frames of labeled LiDAR point clouds in the training set and conduct few-shot fine-tuning experiments. Results are shown in Table \ref{table: nuscenes few-shot results}. Our proposed method \methodshorthand\ achieves 2.11\% mAP and 1.46\% NDS improvement over randomly initialization at convergence, which is the best among all the baselines. When compared to previous SOTA 3D pre-training method UniPAD, \methodshorthand\ achieves $91\%$ more improvement for mAP and $94\%$ more improvement for NDS. If we look into detailed categories, \methodshorthand\ achieves general improvement on different categories. For Car, Barrier, Motorcycle, Pedestrian and Traffic Cone, the improvement are more than 2\% AP. For Bus, \methodshorthand\ introduce an improvement of 5\% AP.

\begin{figure}[t]
\vspace{-2mm}
    \centering
    \includegraphics[width=0.82\linewidth]{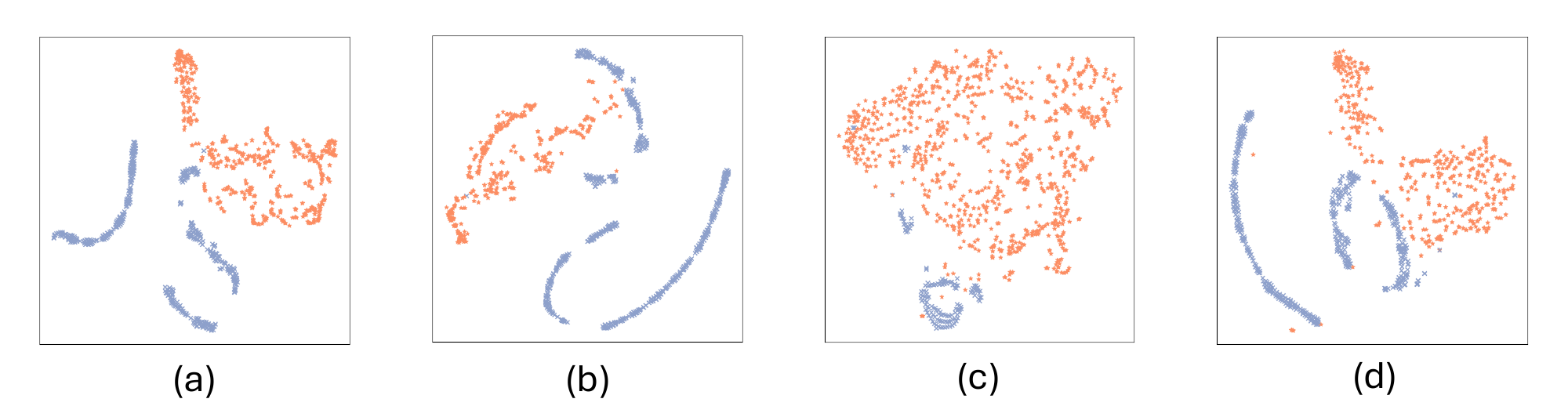}
    \caption{T-SNE visualization of TREND’s features with Moving Object labels. The orange ones are static points while grey blue ones are moving points.}
    \vspace{-2mm}
    \label{fig: T-SNE results}
\end{figure}

\begin{table}[t]
\tablestyle{6pt}{1.1}
\begin{floatrow}
\capbtabbox{
\begin{tabular}{ccc}
\hline
Init.   & mIoU  & Acc   \\ \hline
Rand.   & 28.23 & 70.68 \\
Occ-MAE & 27.82$^{\bf\color{brickred}{-0.41}}$ & 76.57$^{\bf\color{forest}{+5.89}}$ \\
UniPAD  & 29.98 $^{\bf\color{forest}{+1.75}}$ & 76.10$^{\bf\color{forest}{+5.42}}$ \\
TREND   & 31.12$^{\bf\color{forest}{+2.89}}$ & 79.82$^{\bf\color{forest}{+9.14}}$     \\ \hline
\end{tabular}
}{
\vspace{-1mm}
 \caption{Results on SemanticKITTI \cite{semantic_kitti}.}
 \label{table: semantic segmentation}
}
\capbtabbox{
\begin{tabular}{ccccc}
\hline
Rec. Emb. & N. F. & Tem. L. N. F. & mAP & NDS \\ \hline
     \ding{55}               &      \ding{55}         &   \ding{55}                     &   31.06  &  44.75   \\
      \ding{55}               &      \ding{51}         &   \ding{55}                     &   32.16 & 45.26   \\
    \ding{51}               &      \ding{51}         &   \ding{55}                     &   32.45 &
45.76   \\
              \ding{51}       &        \ding{55}      &       \ding{51}                 &  33.17   & 46.21    \\ \hline
\end{tabular}
}{
\vspace{-1mm}
 \caption{Results for ablation study.}
 \label{table: ablation study}
 \small
}
\end{floatrow}
\vspace{-3mm}
\end{table}

\vspace{-2mm}
\subsection{Other Results}
\label{subsec: other results}
\vspace{-1mm}
\noindent\textbf{Transferring Experiments. }We use the backbone pre-trained on Once dataset to initialize CenterPoint \cite{centerpoint} and fine-tune the detector with 1\% training data of Waymo \cite{waymo}. The results are shown in Table \ref{table: transferring}. It can be found that \methodshorthand\ brings an average gain of 0.77 on mAPs and mAPHs, while UniPAD only achieves comparable performance. This demonstrates that \methodshorthand\ is able to pre-train the backbone on one dataset and then transfer to another dataset for performance improvement. 

\noindent\textbf{Accelerating Convergence. }We use the default training iterations in OpenPCDet \cite{pcdet} to train $5\%$ Once data, which is the convergence acceleration setting and the experiment setting in most of the previous 3D pre-training literature. Results are shown in Table \ref{table: once acceleration}. It can be found that T-MAE and UniPAD accelerate convergence and \methodshorthand\ achieves the best performance.

\noindent\textbf{LiDAR Semantic Segmentation. }Results are show in Table \ref{table: semantic segmentation}. It can be found that \methodshorthand\ achieves best performance among different 3D pre-training baselines and improve the performance by 2.89\% in mIoU and 9.14\% in overall accuracy, demonstrating \methodshorthand's generalization ability across tasks.

\noindent\textbf{Ablation Study. }We conduct ablation study to analyze the contribution of different parts of \methodshorthand and results are shown in Table \ref{table: ablation study}. ``Rec. Emb.'',  ``N. F.'' and ``Tem. L. N. F.'' are respectively for Recurrent Embedding, Neural Field and Temporal LiDAR Neural Field. It can be found that using neural field for reconstruction pre-training brings little improvement and even degrades the NDS score compared to training-from-scratch. Adding Recurrent Embedding scheme with neural field improves the performance both on mAP and NDS, which demonstrates that Recurrent Embedding scheme is able to encode 3D features at different timestamps. Finally, with Temporal LiDAR Neural Field, \methodshorthand\ achieves the best performance both on mAP and NDS, showing that Temporal Neural Field better utilizes the temporal information in LiDAR sequence for unsupervised 3D pre-training.

\noindent\textbf{T-SNE for \methodshorthand's features. } We explored whether TREND's pre-trained features can distinguish between moving and static objects. We applied T-SNE \cite{tsne} on TREND's features together with moving/static labels from \cite{mos}. In Figure \ref{fig: T-SNE results}, where static points are colored with orange and moving ones with grey blue, it can be found that although some noise exists, TREND's features for moving and static objects are generally separable after unsupervised pre-training.

\vspace{-2mm}
\section{Conclusion}
\vspace{-1mm}
\label{sec: conclusion}
In this paper, we propose \methodshorthand\ for unsupervised 3D representation learning via temporal forecasting, addressing the temporal embedding and scene representing challenges. With extensive experiments, we demonstrate that \methodshorthand\ is superior in improving downstream performance compared to previous SOTA techniques on various datasets and tasks. These results demonstrate the effectiveness of temporal forecasting in 3D pre-training. We believe \methodshorthand\ will facilitate our understanding on 3D perception in autonomous driving.

\begin{ack}
This paper is supported by the general research fund of Hong Kong 17208825 and 17209324, and the Global Industrial Technology Cooperation Center (GITCC) through a grant agreement with the Korea Institute for Advancement of Technology (KIAT), project number P0028922.
\end{ack}
{
\bibliographystyle{unsrt}
\bibliography{main}

\begin{thebibliography}{10}

\bibitem{second}
Yan Yan, Yuxing Mao, and Bo~Li.
\newblock Second: Sparsely embedded convolutional detection.
\newblock {\em Sensors}, 18(10):3337, 2018.

\bibitem{centerpoint}
Tianwei Yin, Xingyi Zhou, and Philipp Krahenbuhl.
\newblock Center-based 3d object detection and tracking.
\newblock In {\em Proceedings of the IEEE/CVF conference on computer vision and pattern recognition}, pages 11784--11793, 2021.

\bibitem{pv-rcnn}
Shaoshuai Shi, Chaoxu Guo, Li~Jiang, Zhe Wang, Jianping Shi, Xiaogang Wang, and Hongsheng Li.
\newblock Pv-rcnn: Point-voxel feature set abstraction for 3d object detection.
\newblock In {\em Proceedings of the IEEE Conference on Computer Vision and Pattern Recognition}, 2020.

\bibitem{pv-rcnn++}
Shaoshuai Shi, Li~Jiang, Jiajun Deng, Zhe Wang, Chaoxu Guo, Jianping Shi, Xiaogang Wang, and Hongsheng Li.
\newblock Pv-rcnn++: Point-voxel feature set abstraction with local vector representation for 3d object detection.
\newblock {\em arXiv preprint arXiv:2102.00463}, 2021.

\bibitem{sst}
Lue Fan, Ziqi Pang, Tianyuan Zhang, Yu-Xiong Wang, Hang Zhao, Feng Wang, Naiyan Wang, and Zhaoxiang Zhang.
\newblock Embracing single stride 3d object detector with sparse transformer.
\newblock {\em arXiv preprint arXiv:2112.06375}, 2021.

\bibitem{transfusion}
Xuyang Bai, Zeyu Hu, Xinge Zhu, Qingqiu Huang, Yilun Chen, Hongbo Fu, and Chiew-Lan Tai.
\newblock Transfusion: Robust lidar-camera fusion for 3d object detection with transformers.
\newblock In {\em Proceedings of the IEEE/CVF conference on computer vision and pattern recognition}, pages 1090--1099, 2022.

\bibitem{3d_object_detection_survey}
Jiageng Mao, Shaoshuai Shi, Xiaogang Wang, and Hongsheng Li.
\newblock 3d object detection for autonomous driving: A comprehensive survey.
\newblock {\em International Journal of Computer Vision}, 131(8):1909--1963, 2023.

\bibitem{lidar_segmentation_1}
Fangzhou Hong, Hui Zhou, Xinge Zhu, Hongsheng Li, and Ziwei Liu.
\newblock Lidar-based panoptic segmentation via dynamic shifting network.
\newblock In {\em Proceedings of the IEEE/CVF conference on computer vision and pattern recognition}, pages 13090--13099, 2021.

\bibitem{lidar_segmentation_2}
Hui Zhou, Xinge Zhu, Xiao Song, Yuexin Ma, Zhe Wang, Hongsheng Li, and Dahua Lin.
\newblock Cylinder3d: An effective 3d framework for driving-scene lidar semantic segmentation.
\newblock {\em arXiv preprint arXiv:2008.01550}, 2020.

\bibitem{pointcloud_labeling}
Tai Wang, Conghui He, Zhe Wang, Jianping Shi, and Dahua Lin.
\newblock Flava: Find, localize, adjust and verify to annotate lidar-based point clouds.
\newblock In {\em Adjunct Proceedings of the 33rd Annual ACM Symposium on User Interface Software and Technology}, pages 31--33, 2020.

\bibitem{pointcontrast}
Saining Xie, Jiatao Gu, Demi Guo, Charles~R Qi, Leonidas Guibas, and Or~Litany.
\newblock Pointcontrast: Unsupervised pre-training for 3d point cloud understanding.
\newblock In {\em Computer Vision--ECCV 2020: 16th European Conference, Glasgow, UK, August 23--28, 2020, Proceedings, Part III 16}, pages 574--591. Springer, 2020.

\bibitem{contrastive_scene_context}
Ji~Hou, Benjamin Graham, Matthias Nie{\ss}ner, and Saining Xie.
\newblock Exploring data-efficient 3d scene understanding with contrastive scene contexts.
\newblock In {\em Proceedings of the IEEE/CVF conference on computer vision and pattern recognition}, pages 15587--15597, 2021.

\bibitem{P4contrast}
Yunze Liu, Li~Yi, Shanghang Zhang, Qingnan Fan, Thomas Funkhouser, and Hao Dong.
\newblock P4contrast: Contrastive learning with pairs of point-pixel pairs for rgb-d scene understanding.
\newblock {\em arXiv preprint arXiv:2012.13089}, 2020.

\bibitem{strl}
Siyuan Huang, Yichen Xie, Song-Chun Zhu, and Yixin Zhu.
\newblock Spatio-temporal self-supervised representation learning for 3d point clouds.
\newblock In {\em Proceedings of the IEEE/CVF International Conference on Computer Vision}, pages 6535--6545, 2021.

\bibitem{gcc3d}
Hanxue Liang, Chenhan Jiang, Dapeng Feng, Xin Chen, Hang Xu, Xiaodan Liang, Wei Zhang, Zhenguo Li, and Luc Van~Gool.
\newblock Exploring geometry-aware contrast and clustering harmonization for self-supervised 3d object detection.
\newblock In {\em Proceedings of the IEEE/CVF International Conference on Computer Vision}, pages 3293--3302, 2021.

\bibitem{co3}
Runjian Chen, Yao Mu, Runsen Xu, Wenqi Shao, Chenhan Jiang, Hang Xu, Zhenguo Li, and Ping Luo.
\newblock Co\^{} 3: Cooperative unsupervised 3d representation learning for autonomous driving.
\newblock {\em arXiv preprint arXiv:2206.04028}, 2022.

\bibitem{gdmae}
Honghui Yang, Tong He, Jiaheng Liu, Hua Chen, Boxi Wu, Binbin Lin, Xiaofei He, and Wanli Ouyang.
\newblock Gd-mae: generative decoder for mae pre-training on lidar point clouds.
\newblock In {\em Proceedings of the IEEE/CVF Conference on Computer Vision and Pattern Recognition}, pages 9403--9414, 2023.

\bibitem{mvjar}
Runsen Xu, Tai Wang, Wenwei Zhang, Runjian Chen, Jinkun Cao, Jiangmiao Pang, and Dahua Lin.
\newblock Mv-jar: Masked voxel jigsaw and reconstruction for lidar-based self-supervised pre-training.
\newblock In {\em Proceedings of the IEEE/CVF Conference on Computer Vision and Pattern Recognition}, pages 13445--13454, 2023.

\bibitem{ponder}
Di~Huang, Sida Peng, Tong He, Honghui Yang, Xiaowei Zhou, and Wanli Ouyang.
\newblock Ponder: Point cloud pre-training via neural rendering.
\newblock In {\em Proceedings of the IEEE/CVF International Conference on Computer Vision}, pages 16089--16098, 2023.

\bibitem{ponderv2}
Haoyi Zhu, Honghui Yang, Xiaoyang Wu, Di~Huang, Sha Zhang, Xianglong He, Tong He, Hengshuang Zhao, Chunhua Shen, Yu~Qiao, et~al.
\newblock Ponderv2: Pave the way for 3d foundataion model with a universal pre-training paradigm.
\newblock {\em arXiv preprint arXiv:2310.08586}, 2023.

\bibitem{unipad}
Honghui Yang, Sha Zhang, Di~Huang, Xiaoyang Wu, Haoyi Zhu, Tong He, Shixiang Tang, Hengshuang Zhao, Qibo Qiu, Binbin Lin, et~al.
\newblock Unipad: A universal pre-training paradigm for autonomous driving.
\newblock In {\em Proceedings of the IEEE/CVF Conference on Computer Vision and Pattern Recognition}, pages 15238--15250, 2024.

\bibitem{4docc}
Tarasha Khurana, Peiyun Hu, David Held, and Deva Ramanan.
\newblock Point cloud forecasting as a proxy for 4d occupancy forecasting.
\newblock In {\em Proceedings of the IEEE/CVF Conference on Computer Vision and Pattern Recognition}, pages 1116--1124, 2023.

\bibitem{uno}
Ben Agro, Quinlan Sykora, Sergio Casas, Thomas Gilles, and Raquel Urtasun.
\newblock Uno: Unsupervised occupancy fields for perception and forecasting.
\newblock In {\em Proceedings of the IEEE/CVF Conference on Computer Vision and Pattern Recognition}, pages 14487--14496, 2024.

\bibitem{copilot4d}
Lunjun Zhang, Yuwen Xiong, Ze~Yang, Sergio Casas, Rui Hu, and Raquel Urtasun.
\newblock Learning unsupervised world models for autonomous driving via discrete diffusion.
\newblock {\em arXiv preprint arXiv:2311.01017}, 2023.

\bibitem{nerf}
Ben Mildenhall, Pratul~P Srinivasan, Matthew Tancik, Jonathan~T Barron, Ravi Ramamoorthi, and Ren Ng.
\newblock Nerf: Representing scenes as neural radiance fields for view synthesis.
\newblock {\em Communications of the ACM}, 65(1):99--106, 2021.

\bibitem{neus}
Peng Wang, Lingjie Liu, Yuan Liu, Christian Theobalt, Taku Komura, and Wenping Wang.
\newblock Neus: Learning neural implicit surfaces by volume rendering for multi-view reconstruction.
\newblock {\em arXiv preprint arXiv:2106.10689}, 2021.

\bibitem{once}
Jiageng Mao, Minzhe Niu, Chenhan Jiang, Hanxue Liang, Jingheng Chen, Xiaodan Liang, Yamin Li, Chaoqiang Ye, Wei Zhang, Zhenguo Li, et~al.
\newblock One million scenes for autonomous driving: Once dataset.
\newblock {\em arXiv preprint arXiv:2106.11037}, 2021.

\bibitem{nuscenes}
Holger Caesar, Varun Bankiti, Alex~H Lang, Sourabh Vora, Venice~Erin Liong, Qiang Xu, Anush Krishnan, Yu~Pan, Giancarlo Baldan, and Oscar Beijbom.
\newblock nuscenes: A multimodal dataset for autonomous driving.
\newblock In {\em Proceedings of the IEEE/CVF Conference on Computer Vision and Pattern Recognition}, pages 11621--11631, 2020.

\bibitem{waymo}
Pei Sun, Henrik Kretzschmar, Xerxes Dotiwalla, Aurelien Chouard, Vijaysai Patnaik, Paul Tsui, James Guo, Yin Zhou, Yuning Chai, Benjamin Caine, et~al.
\newblock Scalability in perception for autonomous driving: Waymo open dataset.
\newblock In {\em Proceedings of the IEEE/CVF Conference on Computer Vision and Pattern Recognition}, pages 2446--2454, 2020.

\bibitem{semantic_kitti}
J.~Behley, M.~Garbade, A.~Milioto, J.~Quenzel, S.~Behnke, C.~Stachniss, and J.~Gall.
\newblock {SemanticKITTI: A Dataset for Semantic Scene Understanding of LiDAR Sequences}.
\newblock In {\em Proc.~of the IEEE International Conf. on Computer Vision (ICCV)}, 2019.

\bibitem{adpt}
Jiakang Yuan, Bo~Zhang, Xiangchao Yan, Botian Shi, Tao Chen, Yikang Li, and Yu~Qiao.
\newblock Ad-pt: Autonomous driving pre-training with large-scale point cloud dataset.
\newblock {\em Advances in Neural Information Processing Systems}, 36, 2024.

\bibitem{spot}
Xiangchao Yan, Runjian Chen, Bo~Zhang, Jiakang Yuan, Xinyu Cai, Botian Shi, Wenqi Shao, Junchi Yan, Ping Luo, and Yu~Qiao.
\newblock Spot: Scalable 3d pre-training via occupancy prediction for autonomous driving.
\newblock {\em arXiv preprint arXiv:2309.10527}, 2023.

\bibitem{strl2}
Yanhao Wu, Tong Zhang, Wei Ke, Sabine S{\"u}sstrunk, and Mathieu Salzmann.
\newblock Spatiotemporal self-supervised learning for point clouds in the wild.
\newblock In {\em Proceedings of the IEEE/CVF Conference on Computer Vision and Pattern Recognition}, pages 5251--5260, 2023.

\bibitem{triangle_contrast}
Bo~Pang, Hongchi Xia, and Cewu Lu.
\newblock Unsupervised 3d point cloud representation learning by triangle constrained contrast for autonomous driving.
\newblock In {\em Proceedings of the IEEE/CVF Conference on Computer Vision and Pattern Recognition}, pages 5229--5239, 2023.

\bibitem{tmae}
Weijie Wei, Fatemeh~Karimi Nejadasl, Theo Gevers, and Martin~R Oswald.
\newblock T-mae: Temporal masked autoencoders for point cloud representation learning.
\newblock {\em arXiv preprint arXiv:2312.10217}, 2023.

\bibitem{vidar}
Zetong Yang, Li~Chen, Yanan Sun, and Hongyang Li.
\newblock Visual point cloud forecasting enables scalable autonomous driving.
\newblock In {\em Proceedings of the IEEE/CVF Conference on Computer Vision and Pattern Recognition}, pages 14673--14684, 2024.

\bibitem{slidr}
Corentin Sautier, Gilles Puy, Spyros Gidaris, Alexandre Boulch, Andrei Bursuc, and Renaud Marlet.
\newblock Image-to-lidar self-supervised distillation for autonomous driving data.
\newblock In {\em Proceedings of the IEEE/CVF Conference on Computer Vision and Pattern Recognition}, pages 9891--9901, 2022.

\bibitem{limoe}
Xiang Xu, Lingdong Kong, Hui Shuai, Liang Pan, Ziwei Liu, and Qingshan Liu.
\newblock Limoe: Mixture of lidar representation learners from automotive scenes.
\newblock In {\em Proceedings of the Computer Vision and Pattern Recognition Conference}, pages 27368--27379, 2025.

\bibitem{sonata}
Xiaoyang Wu, Daniel DeTone, Duncan Frost, Tianwei Shen, Chris Xie, Nan Yang, Jakob Engel, Richard Newcombe, Hengshuang Zhao, and Julian Straub.
\newblock Sonata: Self-supervised learning of reliable point representations.
\newblock In {\em Proceedings of the Computer Vision and Pattern Recognition Conference}, pages 22193--22204, 2025.

\bibitem{iae}
Siming Yan, Zhenpei Yang, Haoxiang Li, Chen Song, Li~Guan, Hao Kang, Gang Hua, and Qixing Huang.
\newblock Implicit autoencoder for point-cloud self-supervised representation learning.
\newblock In {\em Proceedings of the IEEE/CVF International Conference on Computer Vision}, pages 14530--14542, 2023.

\bibitem{3d_scene_flow_1}
Christoph Vogel, Konrad Schindler, and Stefan Roth.
\newblock 3d scene flow estimation with a piecewise rigid scene model.
\newblock {\em International Journal of Computer Vision}, 115:1--28, 2015.

\bibitem{3d_scene_flow_2}
Xingyu Liu, Charles~R Qi, and Leonidas~J Guibas.
\newblock Flownet3d: Learning scene flow in 3d point clouds.
\newblock In {\em Proceedings of the IEEE/CVF conference on computer vision and pattern recognition}, pages 529--537, 2019.

\bibitem{3d_scene_flow_3}
Sundar Vedula, Peter Rander, Robert Collins, and Takeo Kanade.
\newblock Three-dimensional scene flow.
\newblock {\em IEEE transactions on pattern analysis and machine intelligence}, 27(3):475--480, 2005.

\bibitem{3d_scene_flow_4}
Zirui Wang, Shuda Li, Henry Howard-Jenkins, Victor Prisacariu, and Min Chen.
\newblock Flownet3d++: Geometric losses for deep scene flow estimation.
\newblock In {\em Proceedings of the IEEE/CVF winter conference on applications of computer vision}, pages 91--98, 2020.

\bibitem{3d_scene_flow_5}
Moritz Menze and Andreas Geiger.
\newblock Object scene flow for autonomous vehicles.
\newblock In {\em Proceedings of the IEEE conference on computer vision and pattern recognition}, pages 3061--3070, 2015.

\bibitem{3d_scene_flow_6}
Ye~Zhang and Chandra Kambhamettu.
\newblock On 3d scene flow and structure estimation.
\newblock In {\em Proceedings of the 2001 IEEE Computer Society Conference on Computer Vision and Pattern Recognition. CVPR 2001}, volume~2, pages II--II. IEEE, 2001.

\bibitem{3d_scene_flow_7}
Qingwen Zhang, Yi~Yang, Peizheng Li, Olov Andersson, and Patric Jensfelt.
\newblock Seflow: A self-supervised scene flow method in autonomous driving.
\newblock In {\em European Conference on Computer Vision}, pages 353--369. Springer, 2024.

\bibitem{3d_scene_flow_8}
Kyle Vedder, Neehar Peri, Ishan Khatri, Siyi Li, Eric Eaton, Mehmet Kocamaz, Yue Wang, Zhiding Yu, Deva Ramanan, and Joachim Pehserl.
\newblock Scene flow as a partial differential equation.
\newblock {\em arXiv e-prints}, pages arXiv--2410, 2024.

\bibitem{3d_scene_flow_9}
Jiuming Liu, Guangming Wang, Weicai Ye, Chaokang Jiang, Jinru Han, Zhe Liu, Guofeng Zhang, Dalong Du, and Hesheng Wang.
\newblock Difflow3d: toward robust uncertainty-aware scene flow estimation with iterative diffusion-based refinement.
\newblock In {\em Proceedings of the IEEE/CVF Conference on Computer Vision and Pattern Recognition}, pages 15109--15119, 2024.

\bibitem{pointrcnn}
Shaoshuai Shi, Xiaogang Wang, and Hongsheng Li.
\newblock Pointrcnn: 3d object proposal generation and detection from point cloud.
\newblock In {\em The IEEE Conference on Computer Vision and Pattern Recognition (CVPR)}, June 2019.

\bibitem{partA2}
Shaoshuai Shi, Zhe Wang, Jianping Shi, Xiaogang Wang, and Hongsheng Li.
\newblock From points to parts: 3d object detection from point cloud with part-aware and part-aggregation network.
\newblock {\em IEEE transactions on pattern analysis and machine intelligence}, 43(8):2647--2664, 2020.

\bibitem{cylinder3d}
Hui Zhou, Xinge Zhu, Xiao Song, Yuexin Ma, Zhe Wang, Hongsheng Li, and Dahua Lin.
\newblock Cylinder3d: An effective 3d framework for driving-scene lidar semantic segmentation.
\newblock {\em arXiv preprint arXiv:2008.01550}, 2020.

\bibitem{pvkd}
Yuenan Hou, Xinge Zhu, Yuexin Ma, Chen~Change Loy, and Yikang Li.
\newblock Point-to-voxel knowledge distillation for lidar semantic segmentation.
\newblock In {\em Proceedings of the IEEE/CVF conference on computer vision and pattern recognition}, pages 8479--8488, 2022.

\bibitem{semantic_segmentation_3}
Hengshuang Zhao, Li~Jiang, Jiaya Jia, Philip~HS Torr, and Vladlen Koltun.
\newblock Point transformer.
\newblock In {\em Proceedings of the IEEE/CVF international conference on computer vision}, pages 16259--16268, 2021.

\bibitem{semantic_segmentation_4}
Xiaoyang Wu, Yixing Lao, Li~Jiang, Xihui Liu, and Hengshuang Zhao.
\newblock Point transformer v2: Grouped vector attention and partition-based pooling.
\newblock {\em Advances in Neural Information Processing Systems}, 35:33330--33342, 2022.

\bibitem{semantic_segmentation_5}
Xiaoyang Wu, Li~Jiang, Peng-Shuai Wang, Zhijian Liu, Xihui Liu, Yu~Qiao, Wanli Ouyang, Tong He, and Hengshuang Zhao.
\newblock Point transformer v3: Simpler faster stronger.
\newblock In {\em Proceedings of the IEEE/CVF Conference on Computer Vision and Pattern Recognition}, pages 4840--4851, 2024.

\bibitem{semantic_segmentation_2}
Xin Lai, Yukang Chen, Fanbin Lu, Jianhui Liu, and Jiaya Jia.
\newblock Spherical transformer for lidar-based 3d recognition.
\newblock In {\em Proceedings of the IEEE/CVF Conference on Computer Vision and Pattern Recognition}, pages 17545--17555, 2023.

\bibitem{positional_encoding_1}
Guolin Ke, Di~He, and Tie-Yan Liu.
\newblock Rethinking positional encoding in language pre-training.
\newblock In {\em International Conference on Learning Representations}, 2021.

\bibitem{positional_encoding_2}
Ashish Vaswani, Noam Shazeer, Niki Parmar, Jakob Uszkoreit, Llion Jones, Aidan~N Gomez, {\L}ukasz Kaiser, and Illia Polosukhin.
\newblock Attention is all you need.
\newblock {\em Advances in neural information processing systems}, 30, 2017.

\bibitem{lidar_nerf_1}
Shengyu Huang, Zan Gojcic, Zian Wang, Francis Williams, Yoni Kasten, Sanja Fidler, Konrad Schindler, and Or~Litany.
\newblock Neural lidar fields for novel view synthesis.
\newblock In {\em Proceedings of the IEEE/CVF International Conference on Computer Vision}, pages 18236--18246, 2023.

\bibitem{lidar_nerf_2}
Zehan Zheng, Fan Lu, Weiyi Xue, Guang Chen, and Changjun Jiang.
\newblock Lidar4d: Dynamic neural fields for novel space-time view lidar synthesis.
\newblock In {\em Proceedings of the IEEE/CVF Conference on Computer Vision and Pattern Recognition}, pages 5145--5154, 2024.

\bibitem{lidar_nerf_3}
Tang Tao, Longfei Gao, Guangrun Wang, Yixing Lao, Peng Chen, Hengshuang Zhao, Dayang Hao, Xiaodan Liang, Mathieu Salzmann, and Kaicheng Yu.
\newblock Lidar-nerf: Novel lidar view synthesis via neural radiance fields.
\newblock {\em arXiv preprint arXiv:2304.10406}, 2023.

\bibitem{sdf_1}
Tony Chan and Wei Zhu.
\newblock Level set based shape prior segmentation.
\newblock In {\em 2005 IEEE Computer Society Conference on Computer Vision and Pattern Recognition (CVPR'05)}, volume~2, pages 1164--1170. IEEE, 2005.

\bibitem{sdf_2}
Ravi Malladi, James~A Sethian, and Baba~C Vemuri.
\newblock Shape modeling with front propagation: A level set approach.
\newblock {\em IEEE transactions on pattern analysis and machine intelligence}, 17(2):158--175, 1995.

\bibitem{pytorch}
Adam Paszke, Sam Gross, Francisco Massa, Adam Lerer, James Bradbury, Gregory Chanan, Trevor Killeen, Zeming Lin, Natalia Gimelshein, Luca Antiga, et~al.
\newblock Pytorch: An imperative style, high-performance deep learning library.
\newblock {\em Advances in neural information processing systems}, 32, 2019.

\bibitem{intensity_prediction}
Han Wang, Chen Wang, and Lihua Xie.
\newblock Intensity-slam: Intensity assisted localization and mapping for large scale environment.
\newblock {\em IEEE Robotics and Automation Letters}, 6(2):1715--1721, 2021.

\bibitem{curriculum_learning_1}
Yoshua Bengio, J{\'e}r{\^o}me Louradour, Ronan Collobert, and Jason Weston.
\newblock Curriculum learning.
\newblock In {\em Proceedings of the 26th annual international conference on machine learning}, pages 41--48, 2009.

\bibitem{curriculum_learning_2}
Xin Wang, Yudong Chen, and Wenwu Zhu.
\newblock A survey on curriculum learning.
\newblock {\em IEEE transactions on pattern analysis and machine intelligence}, 44(9):4555--4576, 2021.

\bibitem{information_bottleneck_1}
Naftali Tishby, Fernando~C Pereira, and William Bialek.
\newblock The information bottleneck method.
\newblock {\em arXiv preprint physics/0004057}, 2000.

\bibitem{information_bottleneck_2}
Alessandro Achille and Stefano Soatto.
\newblock Emergence of invariance and disentanglement in deep representations.
\newblock {\em Journal of Machine Learning Research}, 19(50):1--34, 2018.

\bibitem{information_bottleneck_3}
Yao-Hung~Hubert Tsai, Yue Wu, Ruslan Salakhutdinov, and Louis-Philippe Morency.
\newblock Self-supervised learning from a multi-view perspective.
\newblock {\em arXiv preprint arXiv:2006.05576}, 2020.

\bibitem{information_bottleneck_4}
Haoqing Wang, Xun Guo, Zhi-Hong Deng, and Yan Lu.
\newblock Rethinking minimal sufficient representation in contrastive learning.
\newblock In {\em Proceedings of the IEEE/CVF conference on computer vision and pattern recognition}, pages 16041--16050, 2022.

\bibitem{ALSO}
Alexandre Boulch, Corentin Sautier, Björn Michele, Gilles Puy, and Renaud Marlet.
\newblock {ALSO}: Automotive lidar self-supervision by occupancy estimation.
\newblock In {\em CVPR}, 2023.

\bibitem{occ_mae}
Chen Min, Liang Xiao, Dawei Zhao, Yiming Nie, and Bin Dai.
\newblock Occupancy-mae: Self-supervised pre-training large-scale lidar point clouds with masked occupancy autoencoders.
\newblock {\em IEEE Transactions on Intelligent Vehicles}, 2023.

\bibitem{pcdet}
OpenPCDet~Development Team.
\newblock Openpcdet: An open-source toolbox for 3d object detection from point clouds.
\newblock \url{https://github.com/open-mmlab/OpenPCDet}, 2020.

\bibitem{kaiming_rethinking}
Kaiming He, Ross Girshick, and Piotr Doll{\'a}r.
\newblock Rethinking imagenet pre-training.
\newblock In {\em Proceedings of the IEEE/CVF international conference on computer vision}, pages 4918--4927, 2019.

\bibitem{submanifold_spconv}
Benjamin Graham and Laurens Van~der Maaten.
\newblock Submanifold sparse convolutional networks.
\newblock {\em arXiv preprint arXiv:1706.01307}, 2017.

\bibitem{tsne}
Laurens Van~der Maaten and Geoffrey Hinton.
\newblock Visualizing data using t-sne.
\newblock {\em Journal of machine learning research}, 9(11), 2008.

\bibitem{mos}
Benedikt Mersch, Xieyuanli Chen, Ignacio Vizzo, Lucas Nunes, Jens Behley, and Cyrill Stachniss.
\newblock Receding moving object segmentation in 3d lidar data using sparse 4d convolutions.
\newblock {\em IEEE Robotics and Automation Letters}, 7(3):7503--7510, 2022.

\bibitem{3dgs_1}
Guikun Chen and Wenguan Wang.
\newblock A survey on 3d gaussian splatting.
\newblock {\em arXiv preprint arXiv:2401.03890}, 2024.

\bibitem{3dgs_2}
Tong Wu, Yu-Jie Yuan, Ling-Xiao Zhang, Jie Yang, Yan-Pei Cao, Ling-Qi Yan, and Lin Gao.
\newblock Recent advances in 3d gaussian splatting.
\newblock {\em Computational Visual Media}, 10(4):613--642, 2024.

\bibitem{3dgs_3}
Ben Fei, Jingyi Xu, Rui Zhang, Qingyuan Zhou, Weidong Yang, and Ying He.
\newblock 3d gaussian splatting as new era: A survey.
\newblock {\em IEEE Transactions on Visualization and Computer Graphics}, 2024.

\end{thebibliography}
}


\newpage
\section*{NeurIPS Paper Checklist}

\begin{enumerate}

\item {\bf Claims}
    \item[] Question: Do the main claims made in the abstract and introduction accurately reflect the paper's contributions and scope?
    \item[] Answer: \answerYes{}. 
    \item[] Justification: In Abstract and Section \ref{sec:intro}, we discuss the contributions and scope of the paper.
    \item[] Guidelines:
    \begin{itemize}
        \item The answer NA means that the abstract and introduction do not include the claims made in the paper.
        \item The abstract and/or introduction should clearly state the claims made, including the contributions made in the paper and important assumptions and limitations. A No or NA answer to this question will not be perceived well by the reviewers. 
        \item The claims made should match theoretical and experimental results, and reflect how much the results can be expected to generalize to other settings. 
        \item It is fine to include aspirational goals as motivation as long as it is clear that these goals are not attained by the paper. 
    \end{itemize}

\item {\bf Limitations}
    \item[] Question: Does the paper discuss the limitations of the work performed by the authors?
    \item[] Answer: \answerYes{} 
    \item[] Justification: We discuss the limitation in Appendix \ref{sec: limitations}.
    \item[] Guidelines:
    \begin{itemize}
        \item The answer NA means that the paper has no limitation while the answer No means that the paper has limitations, but those are not discussed in the paper. 
        \item The authors are encouraged to create a separate "Limitations" section in their paper.
        \item The paper should point out any strong assumptions and how robust the results are to violations of these assumptions (e.g., independence assumptions, noiseless settings, model well-specification, asymptotic approximations only holding locally). The authors should reflect on how these assumptions might be violated in practice and what the implications would be.
        \item The authors should reflect on the scope of the claims made, e.g., if the approach was only tested on a few datasets or with a few runs. In general, empirical results often depend on implicit assumptions, which should be articulated.
        \item The authors should reflect on the factors that influence the performance of the approach. For example, a facial recognition algorithm may perform poorly when image resolution is low or images are taken in low lighting. Or a speech-to-text system might not be used reliably to provide closed captions for online lectures because it fails to handle technical jargon.
        \item The authors should discuss the computational efficiency of the proposed algorithms and how they scale with dataset size.
        \item If applicable, the authors should discuss possible limitations of their approach to address problems of privacy and fairness.
        \item While the authors might fear that complete honesty about limitations might be used by reviewers as grounds for rejection, a worse outcome might be that reviewers discover limitations that aren't acknowledged in the paper. The authors should use their best judgment and recognize that individual actions in favor of transparency play an important role in developing norms that preserve the integrity of the community. Reviewers will be specifically instructed to not penalize honesty concerning limitations.
    \end{itemize}

\item {\bf Theory assumptions and proofs}
    \item[] Question: For each theoretical result, does the paper provide the full set of assumptions and a complete (and correct) proof?
    \item[] Answer: \answerNA{} 
    \item[] Justification: There is no theoretical result in this paper.
    \item[] Guidelines:
    \begin{itemize}
        \item The answer NA means that the paper does not include theoretical results. 
        \item All the theorems, formulas, and proofs in the paper should be numbered and cross-referenced.
        \item All assumptions should be clearly stated or referenced in the statement of any theorems.
        \item The proofs can either appear in the main paper or the supplemental material, but if they appear in the supplemental material, the authors are encouraged to provide a short proof sketch to provide intuition. 
        \item Inversely, any informal proof provided in the core of the paper should be complemented by formal proofs provided in appendix or supplemental material.
        \item Theorems and Lemmas that the proof relies upon should be properly referenced. 
    \end{itemize}

    \item {\bf Experimental result reproducibility}
    \item[] Question: Does the paper fully disclose all the information needed to reproduce the main experimental results of the paper to the extent that it affects the main claims and/or conclusions of the paper (regardless of whether the code and data are provided or not)?
    \item[] Answer: \answerYes{}{} 
    \item[] Justification: We provide implementation details in Section \ref{subsec: exp settings} and Appendix \ref{appendix: implementation details}.
    \item[] Guidelines:
    \begin{itemize}
        \item The answer NA means that the paper does not include experiments.
        \item If the paper includes experiments, a No answer to this question will not be perceived well by the reviewers: Making the paper reproducible is important, regardless of whether the code and data are provided or not.
        \item If the contribution is a dataset and/or model, the authors should describe the steps taken to make their results reproducible or verifiable. 
        \item Depending on the contribution, reproducibility can be accomplished in various ways. For example, if the contribution is a novel architecture, describing the architecture fully might suffice, or if the contribution is a specific model and empirical evaluation, it may be necessary to either make it possible for others to replicate the model with the same dataset, or provide access to the model. In general. releasing code and data is often one good way to accomplish this, but reproducibility can also be provided via detailed instructions for how to replicate the results, access to a hosted model (e.g., in the case of a large language model), releasing of a model checkpoint, or other means that are appropriate to the research performed.
        \item While NeurIPS does not require releasing code, the conference does require all submissions to provide some reasonable avenue for reproducibility, which may depend on the nature of the contribution. For example
        \begin{enumerate}
            \item If the contribution is primarily a new algorithm, the paper should make it clear how to reproduce that algorithm.
            \item If the contribution is primarily a new model architecture, the paper should describe the architecture clearly and fully.
            \item If the contribution is a new model (e.g., a large language model), then there should either be a way to access this model for reproducing the results or a way to reproduce the model (e.g., with an open-source dataset or instructions for how to construct the dataset).
            \item We recognize that reproducibility may be tricky in some cases, in which case authors are welcome to describe the particular way they provide for reproducibility. In the case of closed-source models, it may be that access to the model is limited in some way (e.g., to registered users), but it should be possible for other researchers to have some path to reproducing or verifying the results.
        \end{enumerate}
    \end{itemize}

\item {\bf Open access to data and code}
    \item[] Question: Does the paper provide open access to the data and code, with sufficient instructions to faithfully reproduce the main experimental results, as described in supplemental material?
    \item[] Answer: \answerYes{}{} 
    \item[] Justification: We will publish code and models.
    \item[] Guidelines:
    \begin{itemize}
        \item The answer NA means that paper does not include experiments requiring code.
        \item Please see the NeurIPS code and data submission guidelines (\url{https://nips.cc/public/guides/CodeSubmissionPolicy}) for more details.
        \item While we encourage the release of code and data, we understand that this might not be possible, so “No” is an acceptable answer. Papers cannot be rejected simply for not including code, unless this is central to the contribution (e.g., for a new open-source benchmark).
        \item The instructions should contain the exact command and environment needed to run to reproduce the results. See the NeurIPS code and data submission guidelines (\url{https://nips.cc/public/guides/CodeSubmissionPolicy}) for more details.
        \item The authors should provide instructions on data access and preparation, including how to access the raw data, preprocessed data, intermediate data, and generated data, etc.
        \item The authors should provide scripts to reproduce all experimental results for the new proposed method and baselines. If only a subset of experiments are reproducible, they should state which ones are omitted from the script and why.
        \item At submission time, to preserve anonymity, the authors should release anonymized versions (if applicable).
        \item Providing as much information as possible in supplemental material (appended to the paper) is recommended, but including URLs to data and code is permitted.
    \end{itemize}

\item {\bf Experimental setting/details}
    \item[] Question: Does the paper specify all the training and test details (e.g., data splits, hyperparameters, how they were chosen, type of optimizer, etc.) necessary to understand the results?
    \item[] Answer: \answerYes{} 
    \item[] Justification: We discuss the setting in Section \ref{subsec: exp settings}.
    \item[] Guidelines:
    \begin{itemize}
        \item The answer NA means that the paper does not include experiments.
        \item The experimental setting should be presented in the core of the paper to a level of detail that is necessary to appreciate the results and make sense of them.
        \item The full details can be provided either with the code, in appendix, or as supplemental material.
    \end{itemize}

\item {\bf Experiment statistical significance}
    \item[] Question: Does the paper report error bars suitably and correctly defined or other appropriate information about the statistical significance of the experiments?
    \item[] Answer: \answerYes{} 
    \item[] Justification: We discuss it in Section \ref{subsec: main results} and \ref{subsec: other results}.
    \item[] Guidelines:
    \begin{itemize}
        \item The answer NA means that the paper does not include experiments.
        \item The authors should answer "Yes" if the results are accompanied by error bars, confidence intervals, or statistical significance tests, at least for the experiments that support the main claims of the paper.
        \item The factors of variability that the error bars are capturing should be clearly stated (for example, train/test split, initialization, random drawing of some parameter, or overall run with given experimental conditions).
        \item The method for calculating the error bars should be explained (closed form formula, call to a library function, bootstrap, etc.)
        \item The assumptions made should be given (e.g., Normally distributed errors).
        \item It should be clear whether the error bar is the standard deviation or the standard error of the mean.
        \item It is OK to report 1-sigma error bars, but one should state it. The authors should preferably report a 2-sigma error bar than state that they have a 96\% CI, if the hypothesis of Normality of errors is not verified.
        \item For asymmetric distributions, the authors should be careful not to show in tables or figures symmetric error bars that would yield results that are out of range (e.g. negative error rates).
        \item If error bars are reported in tables or plots, The authors should explain in the text how they were calculated and reference the corresponding figures or tables in the text.
    \end{itemize}

\item {\bf Experiments compute resources}
    \item[] Question: For each experiment, does the paper provide sufficient information on the computer resources (type of compute workers, memory, time of execution) needed to reproduce the experiments?
    \item[] Answer: \answerYes{} 
    \item[] Justification: We discuss this in Appendix \ref{appendix: implementation details} and Section \ref{discussions}.
    \item[] Guidelines:
    \begin{itemize}
        \item The answer NA means that the paper does not include experiments.
        \item The paper should indicate the type of compute workers CPU or GPU, internal cluster, or cloud provider, including relevant memory and storage.
        \item The paper should provide the amount of compute required for each of the individual experimental runs as well as estimate the total compute. 
        \item The paper should disclose whether the full research project required more compute than the experiments reported in the paper (e.g., preliminary or failed experiments that didn't make it into the paper). 
    \end{itemize}
    
\item {\bf Code of ethics}
    \item[] Question: Does the research conducted in the paper conform, in every respect, with the NeurIPS Code of Ethics \url{https://neurips.cc/public/EthicsGuidelines}?
    \item[] Answer: \answerYes{} 
    \item[] Justification: 
    \item[] Guidelines:
    \begin{itemize}
        \item The answer NA means that the authors have not reviewed the NeurIPS Code of Ethics.
        \item If the authors answer No, they should explain the special circumstances that require a deviation from the Code of Ethics.
        \item The authors should make sure to preserve anonymity (e.g., if there is a special consideration due to laws or regulations in their jurisdiction).
    \end{itemize}

\item {\bf Broader impacts}
    \item[] Question: Does the paper discuss both potential positive societal impacts and negative societal impacts of the work performed?
    \item[] Answer: \answerYes{} 
    \item[] Justification: We discuss broader impacts in Appendix \ref{appendix: Broader Impact}.
    \item[] Guidelines:
    \begin{itemize}
        \item The answer NA means that there is no societal impact of the work performed.
        \item If the authors answer NA or No, they should explain why their work has no societal impact or why the paper does not address societal impact.
        \item Examples of negative societal impacts include potential malicious or unintended uses (e.g., disinformation, generating fake profiles, surveillance), fairness considerations (e.g., deployment of technologies that could make decisions that unfairly impact specific groups), privacy considerations, and security considerations.
        \item The conference expects that many papers will be foundational research and not tied to particular applications, let alone deployments. However, if there is a direct path to any negative applications, the authors should point it out. For example, it is legitimate to point out that an improvement in the quality of generative models could be used to generate deepfakes for disinformation. On the other hand, it is not needed to point out that a generic algorithm for optimizing neural networks could enable people to train models that generate Deepfakes faster.
        \item The authors should consider possible harms that could arise when the technology is being used as intended and functioning correctly, harms that could arise when the technology is being used as intended but gives incorrect results, and harms following from (intentional or unintentional) misuse of the technology.
        \item If there are negative societal impacts, the authors could also discuss possible mitigation strategies (e.g., gated release of models, providing defenses in addition to attacks, mechanisms for monitoring misuse, mechanisms to monitor how a system learns from feedback over time, improving the efficiency and accessibility of ML).
    \end{itemize}
    
\item {\bf Safeguards}
    \item[] Question: Does the paper describe safeguards that have been put in place for responsible release of data or models that have a high risk for misuse (e.g., pretrained language models, image generators, or scraped datasets)?
    \item[] Answer: \answerNA{} 
    \item[] Justification: 
    \item[] Guidelines:
    \begin{itemize}
        \item The answer NA means that the paper poses no such risks.
        \item Released models that have a high risk for misuse or dual-use should be released with necessary safeguards to allow for controlled use of the model, for example by requiring that users adhere to usage guidelines or restrictions to access the model or implementing safety filters. 
        \item Datasets that have been scraped from the Internet could pose safety risks. The authors should describe how they avoided releasing unsafe images.
        \item We recognize that providing effective safeguards is challenging, and many papers do not require this, but we encourage authors to take this into account and make a best faith effort.
    \end{itemize}

\item {\bf Licenses for existing assets}
    \item[] Question: Are the creators or original owners of assets (e.g., code, data, models), used in the paper, properly credited and are the license and terms of use explicitly mentioned and properly respected?
    \item[] Answer: \answerYes{} 
    \item[] Justification: We cite the related papers.
    \item[] Guidelines:
    \begin{itemize}
        \item The answer NA means that the paper does not use existing assets.
        \item The authors should cite the original paper that produced the code package or dataset.
        \item The authors should state which version of the asset is used and, if possible, include a URL.
        \item The name of the license (e.g., CC-BY 4.0) should be included for each asset.
        \item For scraped data from a particular source (e.g., website), the copyright and terms of service of that source should be provided.
        \item If assets are released, the license, copyright information, and terms of use in the package should be provided. For popular datasets, \url{paperswithcode.com/datasets} has curated licenses for some datasets. Their licensing guide can help determine the license of a dataset.
        \item For existing datasets that are re-packaged, both the original license and the license of the derived asset (if it has changed) should be provided.
        \item If this information is not available online, the authors are encouraged to reach out to the asset's creators.
    \end{itemize}

\item {\bf New assets}
    \item[] Question: Are new assets introduced in the paper well documented and is the documentation provided alongside the assets?
    \item[] Answer: \answerNA{} 
    \item[] Justification: 
    \item[] Guidelines:
    \begin{itemize}
        \item The answer NA means that the paper does not release new assets.
        \item Researchers should communicate the details of the dataset/code/model as part of their submissions via structured templates. This includes details about training, license, limitations, etc. 
        \item The paper should discuss whether and how consent was obtained from people whose asset is used.
        \item At submission time, remember to anonymize your assets (if applicable). You can either create an anonymized URL or include an anonymized zip file.
    \end{itemize}

\item {\bf Crowdsourcing and research with human subjects}
    \item[] Question: For crowdsourcing experiments and research with human subjects, does the paper include the full text of instructions given to participants and screenshots, if applicable, as well as details about compensation (if any)? 
    \item[] Answer: \answerNA{} 
    \item[] Justification: 
    \item[] Guidelines:
    \begin{itemize}
        \item The answer NA means that the paper does not involve crowdsourcing nor research with human subjects.
        \item Including this information in the supplemental material is fine, but if the main contribution of the paper involves human subjects, then as much detail as possible should be included in the main paper. 
        \item According to the NeurIPS Code of Ethics, workers involved in data collection, curation, or other labor should be paid at least the minimum wage in the country of the data collector. 
    \end{itemize}

\item {\bf Institutional review board (IRB) approvals or equivalent for research with human subjects}
    \item[] Question: Does the paper describe potential risks incurred by study participants, whether such risks were disclosed to the subjects, and whether Institutional Review Board (IRB) approvals (or an equivalent approval/review based on the requirements of your country or institution) were obtained?
    \item[] Answer: \answerNA{} 
    \item[] Justification: 
    \item[] Guidelines:
    \begin{itemize}
        \item The answer NA means that the paper does not involve crowdsourcing nor research with human subjects.
        \item Depending on the country in which research is conducted, IRB approval (or equivalent) may be required for any human subjects research. If you obtained IRB approval, you should clearly state this in the paper. 
        \item We recognize that the procedures for this may vary significantly between institutions and locations, and we expect authors to adhere to the NeurIPS Code of Ethics and the guidelines for their institution. 
        \item For initial submissions, do not include any information that would break anonymity (if applicable), such as the institution conducting the review.
    \end{itemize}

\item {\bf Declaration of LLM usage}
    \item[] Question: Does the paper describe the usage of LLMs if it is an important, original, or non-standard component of the core methods in this research? Note that if the LLM is used only for writing, editing, or formatting purposes and does not impact the core methodology, scientific rigorousness, or originality of the research, declaration is not required.
    \item[] Answer: \answerNA{} 
    \item[] Justification: 
    \item[] Guidelines:
    \begin{itemize}
        \item The answer NA means that the core method development in this research does not involve LLMs as any important, original, or non-standard components.
        \item Please refer to our LLM policy (\url{https://neurips.cc/Conferences/2025/LLM}) for what should or should not be described.
    \end{itemize}

\end{enumerate}

\newpage

\appendix
\section{More Implementation Details.}
\label{appendix: implementation details}
During pre-training of \methodshorthand, we set the curriculum learning epoch as $N_{\text{curri}}^{1}=12$ and $N_{\text{curri}}^{2}=36$. We use mask augmentation for \methodshorthand\ with a masking rate of 0.9. The ego-motion (action) information is directly computed from the ego-vehicle poses provided in standard autonomous driving datasets (from IMU or GPS). All experiments are implemented with Pytorch framework. All pre-trainings are conducted on 8 A100 GPUs with batch size equals to 3 per GPU. All downstream tasks are trained on 4 A100 GPUs with default settings in OpenPCDet \cite{pcdet} except for training iterations. We will release code and pre-trained models.

\section{Repeated Evaluation.}\label{appendix: repeated experiments}
We use the same random seed in all experiments in the main paper for reproducibility. As repeated evaluation can further reveal the training robustness, we further repeat the experiment on Once ($20\%$ downstream data) for 5 times and compute the mean and standard deviation of the results for randomly initialization, UniPAD and \methodshorthand, which are shown in Table \ref{table: once repeated evaluation}. It can be found that \methodshorthand\ still achieves the best performance in mAP while largely reducing the standard deviation. This means pre-training by \methodshorthand\ alleviates the influence of random seed and makes training more stable.

\begin{table}[h]
\centering
\renewcommand\arraystretch{1.1}
\caption{Results for repeated evaluation on Once dataset \cite{once} with $20\%$ downstream data. Mean and variance are in \%.}
\label{table: once repeated evaluation}
\vspace{2mm}
\begin{tabular}{ccccc}
\hline
Init. & mAP & Vehicle & Pedestrian & Cyclist \\ \hline
 Rand.     &  57.29$\pm$0.29   & 68.99$\pm$0.10    & 43.29$\pm$0.93    & 59.59$\pm$0.23     \\
UniPAD     &   58.00$\pm$0.32  &  71.78$\pm$0.16    & 41.78$\pm$0.70     &  60.44$\pm$0.48    \\
 \methodshorthand     &  58.74$\pm$0.11  &  73.07$\pm$0.16    & 42.02$\pm$0.39     & 61.12$\pm$0.22   \\ \hline
\end{tabular}
\end{table}

\section{More Experiments on NuScenes}\label{appendix: fine-tune on nuscenes}
In this section, we conduct more fine-tuning experiments on NuScenes dataset. Specifically, we randomly sample 2.5\% and 5\% of NuScenes training set and train the randomly initialization model \cite{transfusion} until convergence is observed. Then we apply the pre-trained weight by \methodshorthand\ to initialize the model \cite{transfusion} and fine-tune it with the same training iterations. Results are shown in Table \ref{table: more results on nuscenes}. It can be found that \methodshorthand\ consistently improve the performance in downstream 3D object detection task with different ratio of downstream training data.

\begin{table}[h]
\centering
\setlength{\tabcolsep}{3.4pt}
\renewcommand\arraystretch{1.3}
\begin{footnotesize}
\begin{tabular}{c|c|c|c|cccccccc}
\hline
Init. & F.T.                   & mAP   & NDS   & Car   & Truck & Bus   & Barrier & Mot.  & Bic.  & Ped.  & T.C.  \\ \hline
Rand. & \multirow{2}{*}{2.5\%} & 45.35 & 55.36 & 76.74 & 40.89 & 50.07 & 57.48   & 41.58 & 26.13 & 76.67 & 55.77 \\
\methodshorthand &                        & 45.79$^{\bf\color{forest}{+0.64}}$ & 56.23$^{\bf\color{forest}{+0.87}}$ & 77.74 & 42.96 & 50.78 & 59.39   & 40.37 & 23.48 & 77.22 & 57.51 \\ \hline
Rand.  & \multirow{2}{*}{5\%}   & 51.56 & 60.24 & 80.22 & 48.56 & 58.69 & 63.42   & 50.84 & 36.59 & 79.29 & 60.30  \\
\methodshorthand &                        & 52.02$^{\bf\color{forest}{+0.46}}$ & 61.02$^{\bf\color{forest}{+0.78}}$ & 80.54 & 48.15 & 57.93 & 63.57   & 52.59 & 36.92 & 79.99 & 60.94 \\ \hline
\end{tabular}
\end{footnotesize}
\caption{Results for few shot fine-tuning on NuScenes \cite{nuscenes} dataset. We randomly sample 2.5\% and 5\% of labeled point clouds in the training set and use Transfusion \cite{transfusion} as the downstream model for all the experiments here. Results of overall performance (mAP) and different categories (APs) are provided. ``Init.'' indicates the initialization methods. ``Rand'' indicates the results where we gradually increase training iterations for train-from-scratch model until convergence is observed. Mot., Bic., Ped. and T.C. are abbreviations for Motorcycle, Bicycle, Pedestrian and Traffic Cone. We use green color to highlight the performance improvement brought by different initialization methods and bold fonts for best performance in mAP and NDS. All the results are in \%.}
\label{table: more results on nuscenes}
\end{table}

\begin{figure}[ht]
    \centering
    \includegraphics[width=0.75\linewidth]{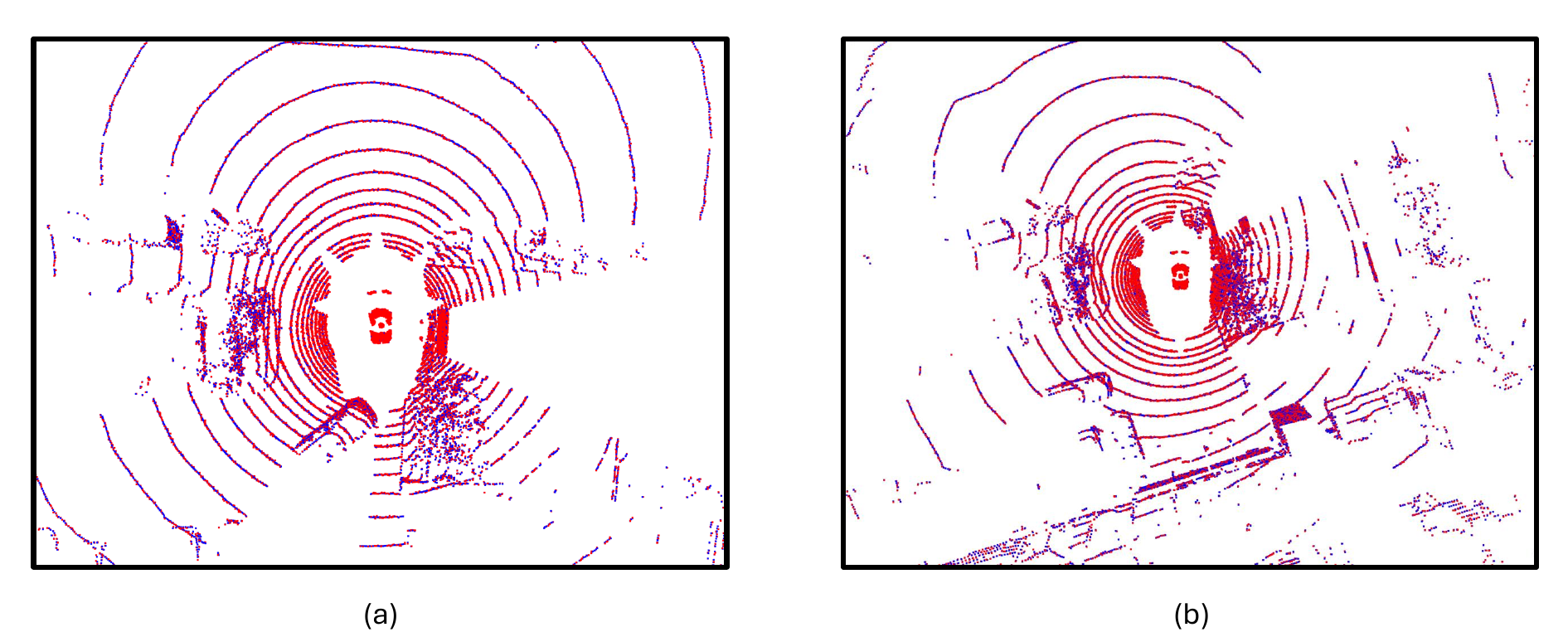}
    \caption{Forecasting Results. Blue points are raw observation from the dataset and red ones are prediction from \methodshorthand.}
    \label{fig: forecasting results}
\end{figure}

\section{Pre-training Decoder Choice Discussion. }
\label{appendix: pre-training decoder choice}

There exists several other decoder choice including occupancy decoder \cite{4docc, spot} and 3DGS (gaussian splatting) \cite{3dgs_1,3dgs_2,3dgs_3}. However, we consider neural-field decoder a more reasonable choice because the rendering process actively involve empty space into pre-training, of which information is actually crucial in LiDAR perception: a) In 3D perception, not only does the detectors need to detect where the objects are but also need to predict characteristics of objects (category, size, velocity and so on). Only use occupied space will harm the performance of 3D detectors. b) Almost all of the current SOTA 3D detectors \cite{second, centerpoint, pv-rcnn, pv-rcnn++, sst, transfusion, 3d_object_detection_survey} first generate a dense BEV-view feature map and predict the bounding boxes and object characteristics based on the dense feature map, which describes both occupied and empty space. c) Neural fields models signed distance values and model the entire scene including occupied and empty space, capturing the relationship between points and their surrounding space. This is significant for understanding scene geometry and potential object trajectories, providing a complete understanding of the environment. We further conduct experiments with 3DGS, occupancy, Copilot4D \cite{copilot4d} and ViDAR \cite{vidar} decoder on NuScenes. Downstream results of mAPs are 31.06\% (random init), 30.84\% (occupancy), 31.90\% (GS-TREND), 31.08\% (Copilot4D), 30.76\% (ViDAR) and 33.17\% (TREND). It can be found that replacing neural field in TREND with 3DGS or occupancy decoder degrades the representation quality.



\section{More Visualizations}\label{appendix: visualizations}

\textbf{Forecasting results of \methodshorthand. }Qualitative visualization would enhance understanding of how TREND learns object motions. We first generate a visualization of forecasting results from \methodshorthand\ in Figure \ref{fig: forecasting results}, where blue points are observation from dataset and red ones are prediction from TREND. As the forecasting error is small, it can be found that the difference is hard to be observed on the figure. Furthermore, we compute Mean Square Error on range prediction. Furthermore, we compute this error for moving objects and static objects respectively. Results (in meters) are: 0.0140. It can be found that the error is in centi-meter scale.

\section{Experiments on Hyper-parameter Sensitivity and Curriculum Learning.}
\label{appendix: Experiments on Hyper-parameter Sensitivity and Curriculum Learning}

\textbf{Forecasting Length. }In our experiments, we used a maximum forecast horizon of 4 frames (approximately 2 seconds). We conduct experiments on shorter and longer horizons on Once with 100\% downstream labels. Results are 65.65\% (3 frames), 66.09\% (4 frames) and 65.33\% (5 frames). We observe that representation quality first increases and then decreases when we add more timestamps for forecasting. First of all, it demonstrates that temporal information helps representation learning. Then, as longer sequence are not as predictable as shorter ones, the representation quality degrades after 2 seconds.

\textbf{Curriculum Learning Strategy. }The curriculum learning strategy is indeed important for TREND's performance. We conducted an ablation experiment training TREND without curriculum learning on Once dataset with 100\% downstream labels. Result of mAP is 65.34\%. It show that without curriculum learning, performance of TREND drops by 0.75\%, demonstrating that gradually increasing forecasting complexity is crucial for effective representation learning.

\textbf{Masking Rate. }The 90 \% masking rate is determined empirically. We further conduct sensitivity study on masking rates on Nuscenes. Results of mAP are 32.56\% for 80 percent masking rate, 33.17\% for 90 percent, and 32.88\% for 95 percent. While masking contributes to performance improvement, the temporal forecasting remains the primary driver of TREND's effectiveness.

\textbf{Sampling Strategy. } When sampling rays for rendering, we first filter ground LiDAR points (most of the ground points are background and less informative in pre-training) using the height of LiDAR sensor, which is originally provided in the datasets. Then we conduct uniform sampling. To investigate the influence of sampling strategy, we conduct ablation study of different sampling strategies including fully uniform sampling, farthest point sampling, uniform sampling with ground points filtering and farthest point sampling with ground points filtering. The experiments are conducted on NuScenes dataset. Results on mAP are 31.65\% for FPS, 31.68\% for uniform sampling, 32.52\% for FPS with ground point filtering and 33.17\% for uniform sampling with ground point filtering. It can be found that different sampling strategies make little difference but filtering ground points matters because ground points are background and less informative for the backbone pre-training.

\section{Comparison to pre-training method leveraging 2D images.}

Leveraging 2D image priors to pre-traing LiDAR encoder also serves as a promising direction. Research efforts include SLiDR \cite{slidr}, LiMoE \cite{limoe} and Sonota \cite{sonata}. We further conduct experiments comparing these methods. As LiMoE(\cite{limoe})'s official repository only publish the first stage of its training, our experiment here utilize this part of code to pre-train the same LiDAR backbone we use. We also apply Sonota \cite{sonata} to pre-train the same LiDAR backbone we use. Experiments are conducted on NuScenes dataset. Results are as belows:

\begin{table}[h]
\begin{tabular}{c|cc}
\hline
Init.        & mAP & NDS \\ \hline
From-scratch &  31.06   &  44.75   \\
LiMoE (first stage)    &  32.21   &  45.61   \\
Sonota       &  32.32   &  46.07   \\
TREND        &  33.17   &  46.21   \\ \hline
\end{tabular}
\end{table}

It can be found that incorporating 2D prior yields similar results (a bit lower ; $<1\%$ difference in mAP and NDS) as TREND, demonstrating the effectiveness of distillation from 2D prior. Meanwhile, as TREND only uses LiDAR modality for pre-training, it can be demonstrated that incorporating temporal information helps learn good 3D representations for downstream perception task. It would be a promising direction to bring both 2D prior and temporal information for pre-training.

\section{Broader Impact}
\label{appendix: Broader Impact}

This paper presents \methodshorthand, an unsupervised 3D representation learning method for LiDAR perception tasks in autonomous driving. There are three potential societal consequences of our work.

\noindent\textbf{Enhanced Safety and Robustness. }As experiment results show, \methodshorthand\ is able to improve performance on different tasks in autonomous driving, which enables autonomous vehicles (AVs) to better understand and adapt to complex environments without relying on extensive labeled datasets. This can lead to improved generalization across diverse road conditions, reducing the risk of accidents caused by unseen scenarios or edge cases.

\noindent\textbf{Environmental and Economic Benefits. }By reducing the reliance on manually annotated data, \methodshorthand\ lowers the computational and labor costs associated with dataset creation. Also, improved AV perception can lead to more energy-efficient driving behaviors, reducing fuel consumption.

\noindent\textbf{Job Displacement and Workforce Transition.} The adoption of unsupervised 3D pre-training in AVs could accelerate automation in the transportation sector, potentially displacing jobs in trucking, taxi services, and delivery industries.

\section{Limitations}
\label{sec: limitations}
While \methodshorthand\ demonstrates significant improvements over previous unsupervised 3D representation learning methods, two limitations should be acknowledged.

(1) Our approach shows varying effectiveness across different object classes. As observed in our experiments, TREND achieves substantial improvements for vehicle and cyclist classes but on Once dataset shows limited gains for pedestrian detection in low-data regimes. This is likely because pedestrians appear as cylinder-like shapes in LiDAR point clouds, making them less distinguishable from other similar structures in the environment.

(2) Our method currently focuses on the geometric aspects of temporal forecasting without explicitly modeling semantic. Incorporating semantic priors from other sensors like camera could potentially enhance the learned representations.

\end{document}